\documentclass[lettersize,journal]{IEEEtran}
\usepackage{amsmath,amsfonts}
\usepackage{algorithmic}
\usepackage{algorithm}
\usepackage{array}

\usepackage[caption=false,font=footnotesize,labelfont=rm,textfont=rm]{subfig}
\usepackage{textcomp}
\usepackage{stfloats}
\usepackage{url}
\usepackage{verbatim}
\usepackage{graphicx}
\usepackage{cite}
\usepackage{multirow}%
\usepackage{booktabs} 
\usepackage{amsmath,amssymb,amsfonts}
\usepackage{orcidlink}

\begin{document}

\title{Cross-DINO: Cross the Deep MLP and Transformer for Small Object Detection}

\author{
Guiping Cao\orcidlink{0000-0002-0682-2158},
Wenjian Huang\orcidlink{0000-0003-2408-8302},
Xiangyuan Lan\orcidlink{0000-0001-8564-0346},
Jianguo Zhang\orcidlink{0000-0001-9317-0268}, 
\textit{Senior Member, IEEE},\\
Dongmei Jiang\orcidlink{0000-0002-6238-8499},
and Yaowei Wang\orcidlink{0000-0003-2197-9038}, \textit{Member, IEEE}

\thanks{Manuscript received 28 August, 2024; revised 27 December, 2024, and 3 February, 2025; accepted 3 February, 2025. This work was supported in part by the National Key Research and Development Program of China under Grant 2021YFF1200800, in part by the National Natural Science Foundation of China under Grant 62402252, in part by the Shenzhen International Research Cooperation Project under Grant GJHZ20220913142611021, and in part by the Pengcheng Laboratory Research Project under Grants PCL2023A08 and PCL2024Y02. (\textit{Corresponding author: Jianguo Zhang, Xiangyuan Lan}.)}
\thanks{
Guiping Cao and Jianguo Zhang are with the Research Institute of Trustworthy Autonomous Systems and Department of Computer Science and Engineering, Southern University of Science and Technology, Shenzhen 518055, China, and also with the Pengcheng Laboratory (e-mail: 12131099@mail.sustech.edu.cn,  zhangjg@sustech.edu.cn).
}
\thanks{Wenjian Huang is with the Southern University of Science and Technology, Shenzhen 518055, China (e-mail: huangwj@sustech.edu.cn).}
\thanks{Xiangyuan Lan is with the Pengcheng Laboratory, and also with the Pazhou Laboratory (Huangpu) (e-mail: lanxy@pcl.ac.cn).}
\thanks{Dongmei Jiang is with Pengcheng Laboratory (e-mail: jiangdm@pcl.ac.cn).}
\thanks{Yaowei Wang is with the Harbin Institute of Technology at Shenzhen, 
and also with the Pengcheng Laboratory (e-mail: wangyw@pcl.ac.cn).}
\thanks{This paper has supplementary downloadable material available at http://ieeexplore.ieee.org. The file contains appendix content related to the CLAP method, visualizations and analyses of detection results and effective receptive fields, as well as some discussions. The material is 2.4 MB in size.}
}

\markboth{IEEE TRANSACTIONS ON MULTIMEDIA}%
{Shell \MakeLowercase{\textit{et al.}}: Bare Demo of IEEEtran.cls for IEEE Journals}




\maketitle

\begin{abstract}
Small Object Detection (SOD) poses significant challenges due to limited information and the 
model's low class prediction score. 
While Transformer-based detectors have shown promising performance, their potential for SOD remains largely unexplored.
In typical DETR-like frameworks, the CNN backbone network, specialized in aggregating local information, struggles to capture the necessary contextual information for SOD. 
The multiple attention layers in the Transformer Encoder face difficulties in effectively attending to small objects and can also lead to blurring of features.
Furthermore, the model's lower class prediction score of small objects compared to large objects further increases the difficulty of SOD.
To address these challenges, we introduce a novel approach called \textbf{Cross-DINO}. 
This approach incorporates the deep MLP network to aggregate initial 
feature representations with both short and long range information for SOD.
Then, a new Cross Coding Twice Module (CCTM) is applied to integrate these initial representations to the Transformer Encoder feature, enhancing the details of small objects.
Additionally,
we introduce a new kind of soft label named Category-Size (CS), integrating the Category and Size of objects. 
By treating CS as new ground truth, we propose a new loss function called Boost Loss to improve the class prediction score of the model.
Extensive experimental results on COCO, WiderPerson, VisDrone, AI-TOD, and SODA-D datasets demonstrate that Cross-DINO efficiently improves the performance of DETR-like models on SOD. Specifically, our model achieves \textbf{36.4\%} AP$_S$ on COCO for SOD with only 45M parameters, outperforming the DINO by \textbf{+4.4\%} AP$_S$ (36.4\% vs. 32.0\%) with fewer parameters and FLOPs, under 12 epochs training setting. 
The source codes will be available at ~\href{https://github.com/Med-Process/Cross-DINO/tree/main}{https://github.com/Med-Process/Cross-DINO}.
\end{abstract}


\section{Introduction}
\IEEEPARstart{O}{bject} detection (OD) holds a significant importance in the domain of computer vision. 
The goal is to identify the bounding boxes and recognize categories for the objects of interest within images.
Over the years, significant advancements have been made in the field of OD due to the advent of deep learning, particularly the convolutional neural network (CNN)~\cite{ren2015faster, redmon2016you, liu2016ssd, he2017mask, wang2023yolov7}.
Recently, transformer-based detectors, such as \textbf{DE}tection \textbf{TR}ansformer (DETR)~\cite{carion2020end} and \textbf{D}ETR with \textbf{I}mproved de\textbf{N}oising anch\textbf{O}r boxes (DINO)~\cite{zhang2022dino}, are novel \textit{end-to-end} detection frameworks that exhibit promising performance compared to the CNN-based detection methods~\cite{zhang2022dino,zheng2023less}.
However, even though the accurate detection of medium-size and large-size objects has been achieved in many applications, accurate Small Object Detection (SOD) remains an extremely challenging task for diverse object detection methods~\cite{liu2021survey, rekavandi2023transformers}.
In the context of OD, the term ``small object" refers to the objects that occupy a small portion of the input image. 
Specifically, in the widely utilized MS COCO dataset~\cite{lin2014microsoft}, small objects are defined as those with a bounding box size of $32 \times 32$ pixels or smaller in a typical $480 \times 640$ image.

\begin{figure}[t]
 \centering
 \small
     \subfloat[AP on COCO]{
		\begin{minipage}[t]{0.49\linewidth}
			\centering
			\includegraphics[width=0.99\linewidth]{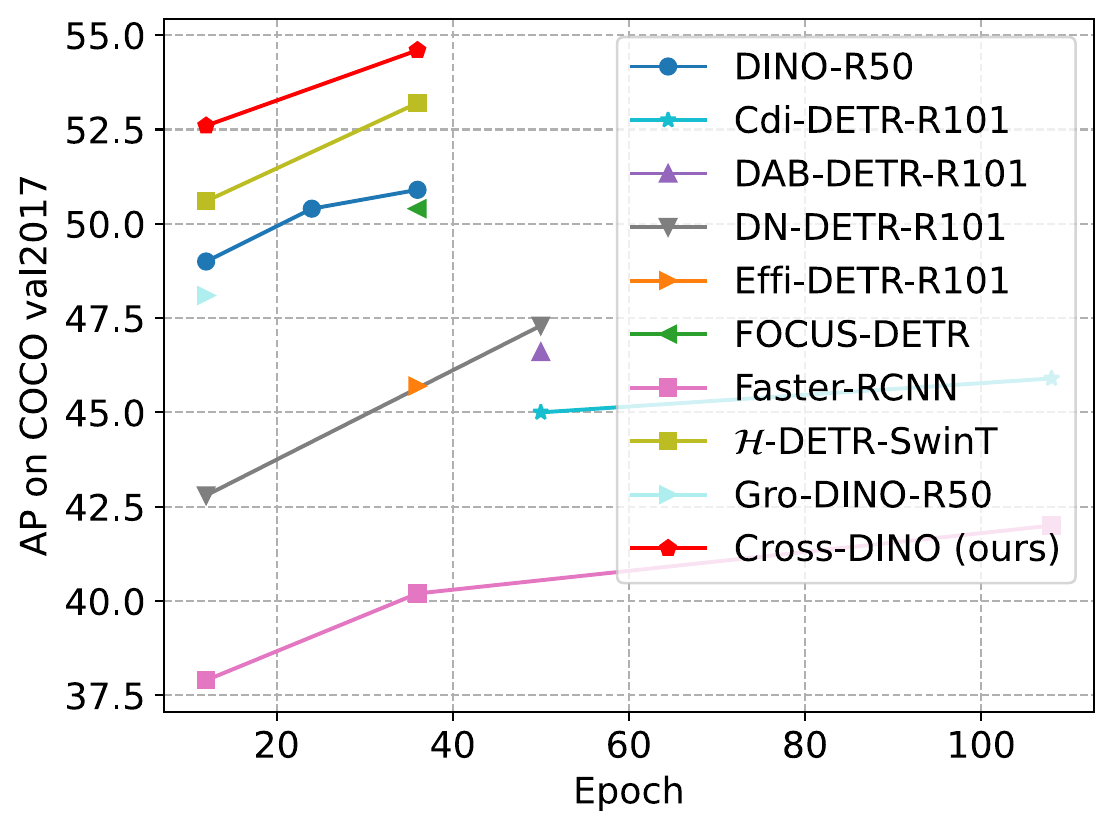}
		\end{minipage}
  }
  \subfloat[AP$_S$ on COCO]{
		\begin{minipage}[t]{0.48\linewidth}
			\centering
			\includegraphics[width=0.99\linewidth]{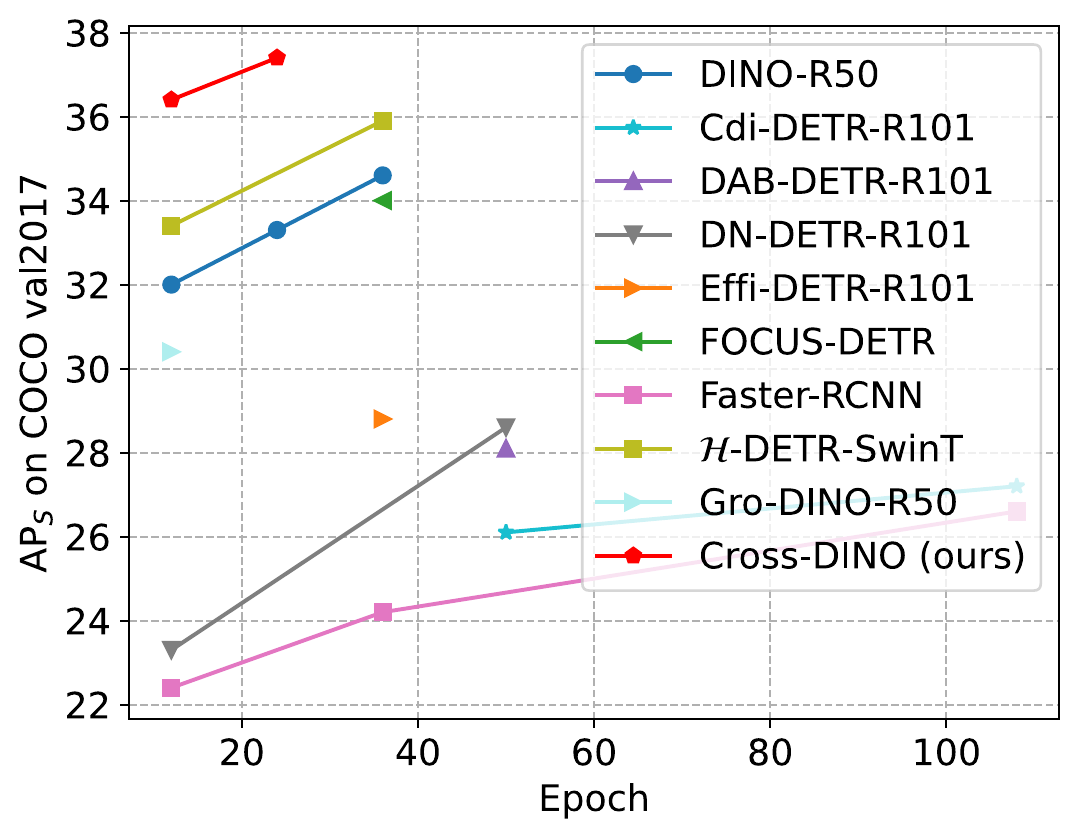}
		\end{minipage}
  }
     \caption{Comparison of different models of AP and AP$_S$ w.r.t the different training epochs on \texttt{val2017} of COCO. When compared to popular CNN-based and transformer-based models, Cross-DINO gets the higher AP$_S$ for SOD and AP for general OD.
      }
    \label{fig:ap}
\end{figure}

\begin{figure}[t]
  \centering
   \includegraphics[width=1.0\linewidth]{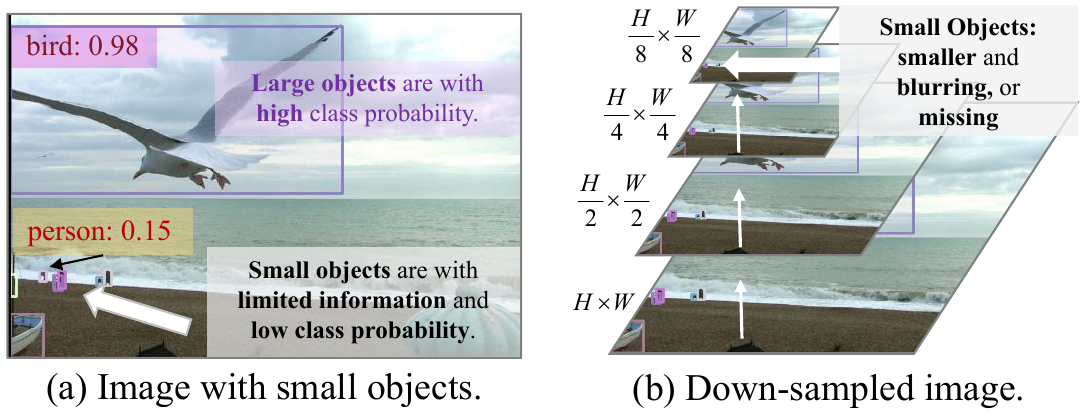}
   \caption{Challenges of SOD. (a) In contrast to large objects, small objects often exhibit \textit{lower} score in class prediction.   
   (b) Objects become smaller when down-sampling occurs. Down-sampling reduces the image's spatial resolution and blurs the image, making it difficult to capture fine details of small objects.}
   \label{fig:so}
\end{figure}

\begin{figure}[t]
  \centering
   \includegraphics[width=0.9\linewidth]{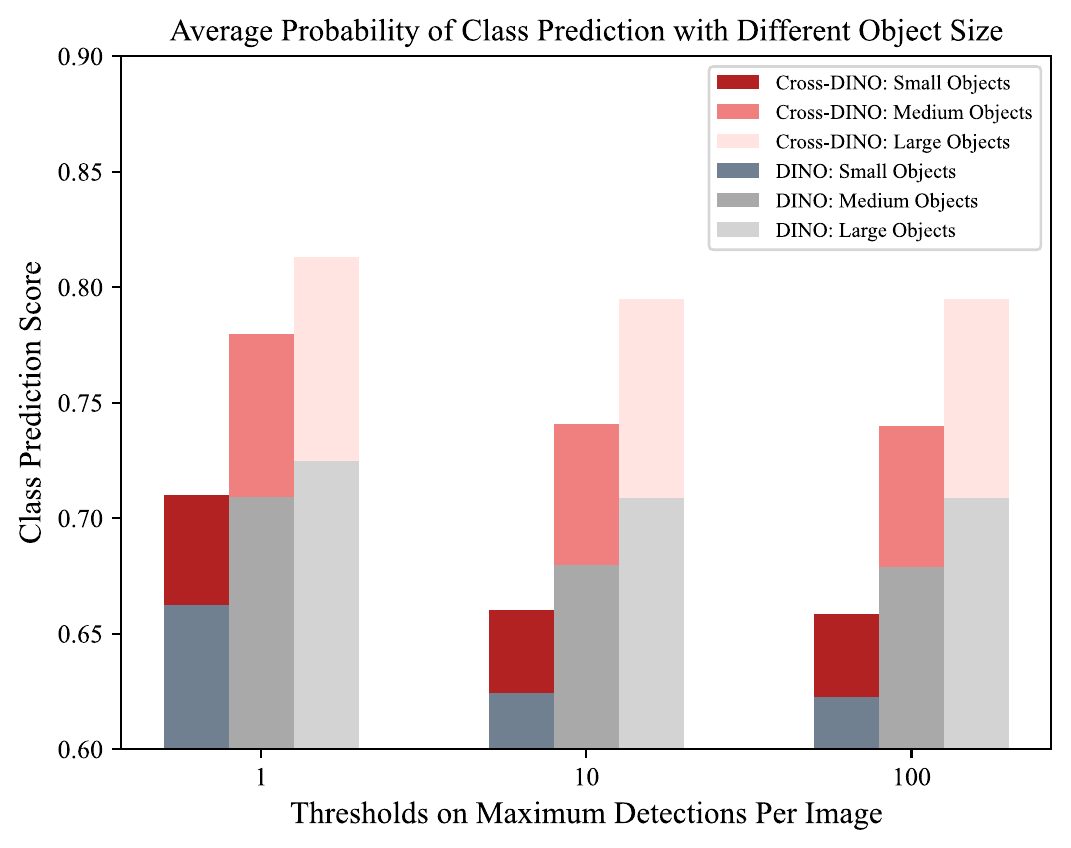}
   \caption{The average class prediction scores of DINO-4scale~\cite{zhang2022dino} and Cross-DINO model with Boost loss under different object size on \texttt{val2017} of COCO~\cite{lin2014microsoft}.
    It's worth noting that Cross-DINO detects additional `harder' small objects than DINO. These `harder' objects are detected with lower scores, resulting in a decrease in the average confidence scores. For a fair comparison, we computed the statistical results for objects with a detection confidence threshold of 0.4 or higher for both models.   
   }
   \label{fig:class_pro}
\end{figure}

\textbf{Challenges of SOD.} SOD is a highly challenging task due to the limited information available for small objects and various factors associated with the model:
\begin{itemize}
    \item \textit{Limited Information}: small objects typically occupy only a few pixels, as illustrated in Fig.~\ref{fig:so} (a), leading to low visibility and making it challenging to obtain detailed information. Additionally, they often have low contrast, indistinct boundaries, or occlusions, further complicating their differentiation from the background or other objects.
    \item \textit{Object Missing and Feature Blurring}: it is challenging for attention layers to pay more attention on small objects, particularly when the features are of lower-resolution, as shown in Fig.~\ref{fig:so} (b). Furthermore, these attention layers would blur the features due to inaccurate attention maps, increasing the difficulties of SOD.
    \item \textit{Low Score in Class Prediction}: Fig.~\ref{fig:class_pro} shows that object detection models usually exhibit \textit{lower} score in classifying small objects than large objects, resulting in a reduced detection capability for small objects.
\end{itemize}

Due to these notable challenges, there still remains a considerable performance gap between SOD and the normal-sized object detection.
These challenges highlight \textit{three} critical factors within SOD models.
Firstly, the limited information poses the need for detection models to aggregate more comprehensive feature representations that capture contextual relationships with neighboring objects for small objects~\cite{liu2021survey}. Secondly, fine-grained features are crucial for small objects to compensate for blurred features. Thirdly, enhancing the class prediction score plays a vital role in achieving accurate SOD.
Despite the success of DETR-like methods in general object detection tasks, they have not adequately addressed the three aforementioned challenges in SOD~\cite{cheng2023towards}.

\textbf{Limitations of DETR-like Models on SOD.} DETR-like models cast the process of OD as a set prediction problem, employing bipartite matching to align predicted objects with ground truth.
The DETR-like architecture comprises three key modules: a compact backbone for feature extraction, a transformer encoder neck for feature enhancement, and a transformer-decoder for predicting boxes and classes. 
These \textit{three} components interconnect in a \textit{cascaded} manner, which amplifies the model's difficulty in SOD:

\begin{itemize}
    \item \textit{Challenges with initial feature representations:} The detection capability of small objects is heavily influenced by the \textit{initial} feature representations extracted from the input images by a generic backbone model. These initial features have a dominant impact on the quality of both encoder and decoder features. Most DETR-like models~\cite{carion2020end, yao2021efficient, liu2021dab, zhang2022dino} usually adopt ResNet~\cite{he2016deep} as the backbone to generate a lower-resolution activation map, which aggregates the feature only in the \textit{short-range} by the convolution layer with small kernels. These short-range feature maps lack contextual information, which poses challenges in locating and recognizing small objects with limited information.

    \item \textit{Challenges with transformer encoder:} It is challenging for DETR-like models to automatically focus on small objects in lower resolution features through attention operations, leading to the missing of small objects. In addition, multiple attention layers of transformer encoder may generate noise attention maps, tending to blur fine details of features layer-by-layer, as illustrated in Fig.~\ref{fig:vis} of the Encoder Feature, which is horrible for SOD.

    \item \textit{Relations between bounding boxes and classification predictions:} These models predict boxes and classification results separately, neglecting the relations between object size and class prediction score. Consequently, this oversight further decreases the performance of SOD. 
\end{itemize}

\textbf{Cross-DINO.} To address these challenges, we propose a novel model called \textit{Cross-DINO}, aimed at bridging the gap between SOD and generic object detection in DETR-like models. Specifically, (1) we introduce a new deep MLP model that efficiently aggregates both \textit{short-range} and \textit{long-range} information simultaneously, providing richer and more comprehensive \textit{initial} feature representations with context cues from small objects and their neighbors; (2) we propose a novel Cross Coding Twice Module that enables the transformer encoder features to gradually learn fine-grained details from the backbone features in two steps, enhancing the model's capability for SOD; (3) we design a new soft label named Category-Size (CS) and introduce a novel Boost loss to strengthen the class prediction score of small objects by efficiently incorporating information about Category and Size of objects; (4) Extensive experiments indicate Cross-DINO remarkably improves the performance of DINO model in SOD. In particular, Cross-DINO achieves 36.4\% AP$_S$ with only 45M parameters, surpassing DINO by \textbf{+4.4\%} AP$_S$ while utilizing \textit{fewer} parameters and FLOPs under a 12-epoch training setting. Fig.~\ref{fig:ap} presents a comparison between our model and other DETR-like and CNN-based detectors, highlighting its significant superiority in both performance and parameter efficiency.

\section{Related Works}\label{sec:related_work}

\subsection{Object Detection} 
Object detection has heavily relied on CNN-based models for an extensive period of time~\cite{zou2023object,tian2024multi}.
These models are capable of learning powerful and high-level features from images. Two main approaches in OD are anchor-based and anchor-free methods.
The anchor-based methods, such as Faster R-CNN~\cite{ren2015faster}, YOLO~\cite{redmon2016you}, RetinaNet~\cite{lin2017focal}, leverage predefined anchor boxes to propose initial object locations and handle objects of different sizes and aspect ratios.
To address the limitations of anchor design sensitivity and imbalance issues, anchor-free methods, such as CenterNet~\cite{duan2019centernet} and CornerNet~\cite{law2018cornernet}, directly predict object locations and sizes without the need for anchor-based reference points. They simplify the detection pipeline and offer more flexibility in handling objects of varying scales and aspect ratios. However, these methods still rely on the handcrafted components such as
Non-Maximum Suppression (NMS) and complex pipelines,
which significantly limits their overall performance~\cite{zhang2022dino}.

\subsection{Small Object Detection}
SOD is a challenging task due to the \textit{limited information} available for small objects, making it difficult for models to perform well. The research progress on small object detection has been systematically introduced in the survey papers~\cite{liu2021survey, cheng2023towards, rekavandi2023transformers}.
Deep CNN architectures extract hierarchy feature maps by pooling and sub-sampling operations, which are extremely risky for small objects, as it gradually diminishes the effective information. To address these challenges, many effective methods have been proposed: improving feature representation for small objects~\cite{fu2017dssd, pang2019efficient, shou2022object}, incorporating context information for small objects~\cite{yu2015multi, zhang2018single}, correcting class imbalance for small objects~\cite{zhang2018single}, increasing training examples for small objects~\cite{singh2018sniper, kisantal2019augmentation, zoph2020learning, bosquet2023full}.

Recently, a Normalized Wasserstein Distance~\cite{wang2021normalized} metric is introduced to address the sensitivity of the IoU metric to location deviations in tiny objects. To address the severe scale-sample imbalance problems in anchor-based and anchor-free label assignment paradigms, the Receptive Field based Label Assignment (RFLA) method is introduced in ~\cite{xu2022rfla} to enhance tiny object detection by utilizing the Gaussian receptive field prior and balancing the learning process for tiny objects. Based on the newly developed large-scale benchmark dataset of SODA~\cite{cheng2023towards}, a two-stage framework called CFINet~\cite{yuan2023small_cfi} is proposed specifically for small object detection. This framework is built on a coarse-to-fine pipeline and incorporates feature imitation learning to address the challenges of insufficient and low-quality training samples, as well as the uncertain predicting regions of interest for SOD. To make the tiny-object-specific features visible and clear for detection, SR-TOD~\cite{cao2025visible} approach addresses this issue by designing a self-reconstructed tiny object detection method, which can be combined with existing detectors and further enhances their performance.

Although these methods have improved the detection accuracy of small objects, there is still a notable gap between SOD and generic object detection.
Compared to CNN-based detectors, end-to-end transformer-based detectors~\cite{carion2020end, zhang2022dino} have simplified the detection process and achieved higher accuracy in generic object detecion. However, detecting small objects remains a challenging task for DETR-like models~\cite{huang2025_dqdetr}, indicating that further exploration in this area is necessary. QueryDet~\cite{yang2022querydet} utilizes a novel cascade sparse query mechanism to accelerate the inference speed of feature-pyramid based object detectors; however, it struggles with high-resolution features for small object detection. Since the fixed number and position of queries in existing DETR-like models are not well-suited for detecting tiny objects in aerial datasets, DQ-DETR~\cite{huang2025_dqdetr} dynamically adjusts the number of object queries to mitigate the imbalance of instances in aerial images. Despite the advancements these methods have made in small object detection, it is essential to optimize DETR-like models beyond query adjustments to enhance their structural effectiveness for detecting small objects. 
To address the challenges of limited information, feature ambiguity, and low confidence scores in SOD, this paper aggregates long-range and short-range information simultaneously to enrich context features. These enhanced features are then integrated into the transformer encoder, providing richer information for decoder queries to generate more accurate detection results for SOD.

\subsection{Transformer-based Detectors} 
Since the DETR~\cite{carion2020end} is introduced, a series of DETR-like models have emerged to promote detection performance from various aspects. These models aim to accelerate the training process of the model, such as Conditional DETR~\cite{meng2021conditional}, DN-DETR~\cite{li2022dn}, DINO~\cite{zhang2022dino}, $\mathcal{H}$-DETR~\cite{jia2023detrs}, Stable-DINO~\cite{liu2023detection}. 
They also optimize the query formulations, such as Efficient DETR~\cite{yao2021efficient}, DAB-DETR~\cite{liu2021dab}, DINO~\cite{zhang2022dino}, and MLP-DINO~\cite{cao2024mlp}.
Redesigning more advanced transformer encoder and decoder architecture is another way to promote detection accuracy, like Deformable-DET~\cite{zhu2020deformable}, CF-DETR~\cite{cao2022cf}.
Additionally, DiffusionDet~\cite{chen2023diffusiondet} designs a new framework that formulates OD as a denoising diffusion process from noisy boxes to object boxes and achieves favorable performance compared to well-established detectors.
However, it is worth noting that these models primarily address general detection scenarios and may not adequately address the specific challenges of information limitation, object missing and feature blurring, and low confidence in SOD. Therefore, there is a pressing need for the development of effective approaches within DETR-like methods to advance SOD.

\begin{figure*}[t]
  \centering
   \includegraphics[width=0.9\linewidth]{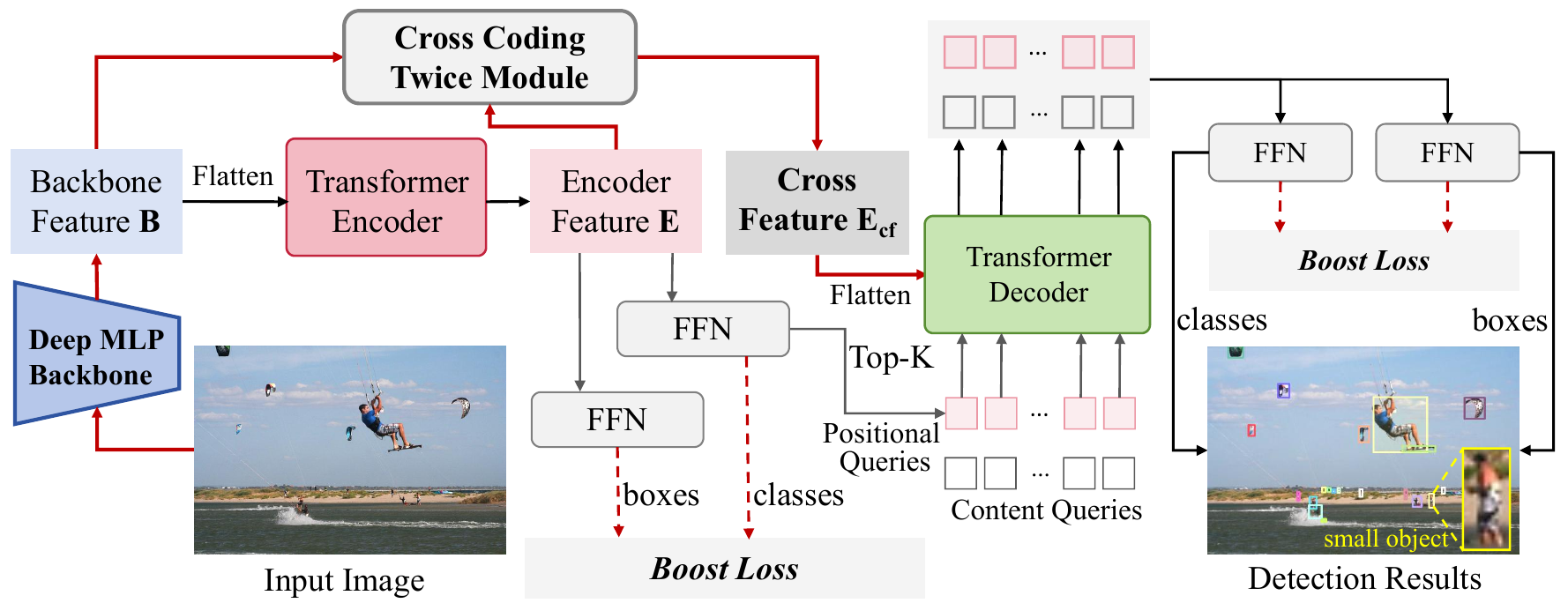}
   \caption{The overall architecture of the proposed Cross-DINO. 
   Our Cross-DINO architecture utilizes a 4-scale backbone feature maps for the decoder to ensure a fair comparison among all models in this study. The red solid and dashed lines highlight the differences between our model and DETR-like models, specifically emphasizing the new deep MLP backbone for enhancing initial representations, the novel CCTM module for strengthening details of objects, and the Boost loss designed to improve detection capabilities, particularly for small objects.}
   \label{fig:cross_dino}
\end{figure*}

\subsection{Diverse Backbone Networks}
In deep learning, the typical backbone networks for vision can be divided into three main categories: CNN-based, Transformer-based, and MLP-based.
CNN is the de-facto standard deep learning network models~\cite{he2016deep, howard2017mobilenets} for vision tasks and has been intensely studied in the vision community. However, they mainly aggregate features \textit{locally} and may struggle with capturing long-term dependencies effectively.
Transformer models~\cite{dosovitskiy2020image, liu2021swin}, on the other hand, introduce attention mechanisms, enabling them to capture long-range information and achieve SOTA accuracy on classification tasks.
One major drawback of transformer-based models is the significant computational burden imposed by the attention mechanism.
Deep MLP~\cite{liu2022we}, a novel architecture primarily based on MLP layers, has shown competitive performance on image classification benchmarks.
Variants such as Strip-MLP~\cite{cao2023strip} presents an efficient and effective token interaction model that simultaneously captures long-range and short-range information while requiring fewer parameters and computational resources, showing great superiority than other popular methods.

Based on the analysis, different backbone networks have varying capabilities in handling small objects and extracting features with contextual information~\cite{liu2021survey}. It is crucial to utilize backbone networks that are applicable to small objects and extract richer feature representations from the limited information of small objects.

\section{Method}\label{sec:method}

In this section, we first describe the overall architecture of Cross-DINO for SOD.
Then, we introduce the deep MLP model into transformer detection framework to aggregate contextual information of small objects in both long-range and short-range.
To address the challenges of feature blurring in the transformer encoder, we propose a new CCTM module, enabling to feed the fine-grained information of backbone feature to encoder feature for small objects.
Finally, we present the novel 
Boost Loss function to alleviate the issue of low class prediction score in SOD.

\subsection{Overall Architecture}

The overall architecture of Cross-DINO is illustrated in Fig.~\ref{fig:cross_dino}, which is extended from the DINO~\cite{zhang2022dino} model for SOD. In particular, Cross-DINO advances in both the model \textit{architecture} and \textit{loss} function to address the challenges of small object detection.

Given the input image, we first introduce the deep MLP backbone model to extract a compact feature representation, referred to as the backbone feature \textit{\textbf{B}}. These features are then flattened and passed through a multi-layer transformer encoder to enhance the features and disentangle the different objects within the image. 
Next, the CCTM is introduced to cross code the backbone feature \textit{\textbf{B}} and the encoder feature \textit{\textbf{E}} to get more fine-grained Cross Feature \textbf{\textit{E}}$_{cf}$.
The flattened Cross Feature \textbf{\textit{E}}$_{cf}$, along with dynamic positional queries and static content queries, is fed into the multi-layer transformer decoder~\cite{zhu2020deformable} for box prediction and refinement step-by-step.
Two feed-forward networks (FFN) are connected to the decoder to predict the class labels and box coordinates.
To address the issue of low class prediction score in small objects, the Boost loss is applied to the encoder feature \textit{\textbf{E}} and all decoder predictions.

\subsection{Cross the Deep MLP and Transformer for SOD}
\label{sec:mlp_trans}

\paragraph{Enhancing the Initial Feature by the Deep MLP Model}

As illustrated in Fig.~\ref{fig:so} of SOD, the limited information available poses a significant challenge for deep learning models in extracting meaningful features.
Many DETR-like frameworks, such as DETR~\cite{carion2020end}, Efficient DETR~\cite{yao2021efficient} and DINO~\cite{zhang2022dino}, commonly utilize ResNet50~\cite{he2016deep} and ResNet101~\cite{he2016deep} as their backbones for feature extraction. However, we argue that the \textit{initial} feature representations extracted from different backbone models have a substantial impact on the subsequent components and overall performance, particularly in detecting small objects.

The ResNet~\cite{he2016deep} model primarily employs convolution layers of small kernels to aggregate information in a \textit{short-range} manner. This approach focuses on a small region of the image while disregarding the contextual information, which is crucial for detecting small objects with limited information.
To address this issue, we explore more efficient vision models and introduce the Strip-MLP~\cite{cao2023strip} into the DINO framework to obtain more comprehensive feature representations. 
Strip-MLP is an MLP-based attention-free model that can simultaneously aggregate information in both \textit{short} and \textit{long} spatial ranges.
This enhances the token interaction power and achieves promising performance in the benchmark of classification, offering advantages in parameters and computational complexity.

\paragraph{Solution to Accept Images of Arbitrary Sizes for the Deep MLP Model}\label{sec:mlp_size}
For deep MLP models, as MLP layers have been operated on the spatial dimension, \textit{the model's weights are related to the size of input images}. This poses a common challenge for most deep MLP models~\cite{liu2022we, cao2023strip} that the model only accepts the \textit{fixed} size of input images.
This limitation restricts their application in downstream dense prediction tasks, such as object detection, where the image input sizes may vary.

To overcome this limitation and incorporate MLP model into the DINO framework, we propose a simple yet effective method called \textbf{C}ropping with \textbf{A}daptive over\textbf{L}a\textbf{P}ping (CLAP) approach that involves adaptively cropping and padding the image into mini-patches and applying the shared weights to all patches. This approach allows us to process images of arbitrary sizes using \textit{fixed feature weights} of deep MLP models, greatly extending their applications on dense prediction tasks.
Further, we incorporate the CLAP approach into Strip-MLP, namely CLAP-Strip-MLP, allowing it to serve as a general vision backbone and plays important roles in downstream tasks.

\subsection{Cross Coding Twice Module}

In DETR-like frameworks, the backbone and transformer encoder play different roles in OD.
To alleviate the computational demands associated with attention operations on input images, the backbone aims to generate \textit{lower-resolution} feature representations \textit{\textbf{B}} for the Transformer encoder. These lower-resolution features allow the subsequent attention modules to process the information more efficiently, reducing the GPU resource requirements and computational workload. The Transformer encoder is designed for enhancing features and disentangling objects, simplifying object extraction and localization for the decoder.

However, when performing multiple layers of attention operations on lower-resolution images, we have observed that it tends to \textit{blur} details in images and \textit{miss} objects, particularly for small objects. 
As shown in Fig.~\ref{fig:vis}, the encoder features are seriously blurred compared to backbone features.
These blurring effects become more pronounced due to the limited information available for small objects, resulting in a significant performance decrease. In contrast, the backbone feature representations contain valuable and fine-grained information that is not present in encoder features. These two types of features with rich detailed information and semantically informative information are both significant for SOD.

\begin{figure}[t]
  \centering
   \includegraphics[width=0.8\linewidth]{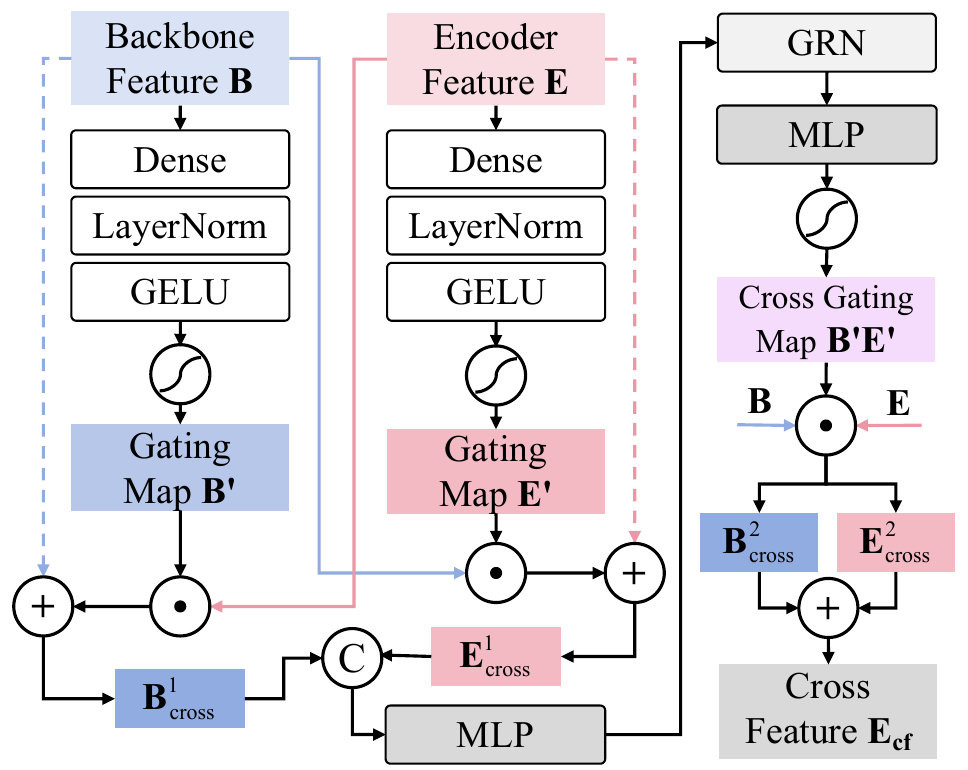}
   \caption{The architecture of Cross Coding Twice Module.}
   \label{fig:cctm}
\end{figure}

Based on the analysis, our motivation is to cross the backbone feature \textit{\textbf{B}} and encoder feature \textit{\textbf{E}} together and re-weight the encoder feature with more fine-grained information for small objects.
Inspired by the commonly used cross-attention~\cite{chen2021crossvit} for selectively gating feature propagation in skip-connections, we propose a new \textbf{C}ross \textbf{C}oding \textbf{T}wice \textbf{M}odule (CCTM) that crosses and codes deep MLP feature \textit{\textbf{B}} and transformer encoder feature \textit{\textbf{E}} in two steps, allowing the feature \textit{\textbf{E}} to \textit{gradually} fuse fine-grained details from the feature \textit{\textbf{B}}, effectively enhancing the performance of SOD.
Fig.~\ref{fig:cctm} presents the architecture of CCTM.
In the first step, we aim to code \textit{crossing} features by mixing these two types of features with gating maps, which can be formulated as (taking \textit{\textbf{E}$_{cross}^{1}$} as an example):
\begin{equation}
  \textbf{\textit{E}}^{'} = \sigma(GELU(LN(FC(\textit{\textbf{E}}))))
  \label{eq:E}
\end{equation}
\begin{equation}
  \textit{\textbf{E}}_{cross}^{1} = \textit{\textbf{E}} + \textit{\textbf{B}} \cdot (1 - \textit{\textbf{E}}^{'})
  \label{eq:first_cctm}
\end{equation}
where $\sigma(*)$ is the sigmoid function. $GELU$~\cite{hendrycks2016gaussian} is the activation layer. $LN$ and $FC$ are the LayerNorm and fully-connected layers, respectively.
The feature \textit{\textbf{E}}$_{cross}^{1} \in \mathbb{R}^{B\times C \times L}$ (where $B,C,L$ represent the batch size, channel number, and token length, respectively) denotes the first cross coding feature of \textit{\textbf{E}} with \textit{\textbf{B}}. Both features \textit{\textbf{E}} and \textit{\textbf{B}} share the same feature dimension as \textit{\textbf{E}}$_{cross}^{1}$.

To increase the contrast and selectivity of channels, we apply Global Response Normalization (GRN)~\cite{woo2023convnext} and two MLP layers to re-weight the feature along the channel dimension.
The second step is designed for adaptively \textit{selecting} the valuable and fine-grained information based on the crossing gating map \textit{\textbf{B}}$'$\textit{\textbf{E}}$'$ from the original feature of \textit{\textbf{B}} and \textit{\textbf{E}}. This process can be formulated as:
\begin{equation}
  \textit{\textbf{E}}_{cf} = 2\textit{\textbf{E}} \cdot \textit{\textbf{B}}'\textit{\textbf{E}}' + \textit{\textbf{B}} \cdot (1 - \textit{\textbf{B}}'\textit{\textbf{E}}')
  \label{eq:second_cctm}
\end{equation}
where \textit{\textbf{E}}$_{cf}$ is the output feature of CCTM. The features \textit{\textbf{E}}$_{cf}$ and \textit{\textbf{B}}$'$\textit{\textbf{E}}$'$ also share the same feature dimension as \textit{\textbf{E}}$_{cross}^1$. To preserve and emphasize the enhanced features, we assign a higher weight of $2$ to the encoder feature.

\subsection{Boost Loss for SOD}
\label{sec:boost}

In SOD, we have observed a common and consistent phenomenon where deep learning-based OD models tend to exhibit \textit{higher} class prediction score of large objects while showing \textit{lower} score on smaller objects. We conduct a statistical analysis on the detection results of the DINO-4scale~\cite{zhang2022dino} model on the \texttt{val2017} of COCO~\cite{lin2014microsoft}. As illustrated in Fig.~\ref{fig:class_pro}, the statistical results indicate that \textit{the average class prediction score for large objects is significantly higher than small objects}.
This finding highlights a strong correlation between object size and the model's score in categories prediction.

However, most existing detection methods~\cite{ren2015faster, he2017mask, zhang2022dino, jia2023detrs} predict the bounding boxes and classifications of the detected objects \textit{independently} and calculate their losses \textit{in isolation}, thereby overlooking the crucial relationships between them.
Based on the aforementioned analysis, to enhance the class prediction score of the model in SOD, we propose a novel loss function, called \textbf{B}lending categ\textbf{O}ry and \textbf{O}bject \textbf{S}ize predic\textbf{T}ion Loss (Boost Loss). 
The motivation of the design is to adaptively increase the classification loss of small objects by incorporating their size into the prediction process to re-weight the probability of the classification. 
This is done by applying a scaling factor to the predicted class probabilities based on the size of the object.

Given an image \textit{\textbf{I}} with size $H \times W$, each object size of the bounding box within the image can be referred as to $h_i \times w_i$ (where $i$ is the index of the box), and the corresponding class label can be denoted as $y_i$. 
Similarly, the object size of the predicted bounding box and classification results of the model for the image can be represented as $\hat{h_i} \times \hat{w_i}$ and $p_i$, respectively. In our Boost loss, we design a novel kind of soft label that incorporates \textbf{C}ategory and \textbf{S}ize of objects (CS), which can be formulated as:
\begin{equation}
  cs_i = \sqrt{\frac{h_i}{H} \times \frac{w_i}{W}} \times y_i
  \label{eq:para}
\end{equation}
Using Eq.~\ref{eq:para}, we can easily construct the new ground truth \textbf{CS} ($cs_{i} \in $ \textbf{CS}) and predictions $\hat{\textbf{CS}}$ based on the ground truth of the dataset and the detection predictions of the model, respectively. However, in some scenes, such as drone images in VisDrone2019~\cite{zhu2021detection} and aerial images in AI-TOD~\cite{wang2021tiny}, the sizes of objects in these images are significantly smaller and the values of $cs$ and $\hat{cs}$ in Boost loss become very small after normalization. For instance, an object with size $8 \times 8$ in a $1024 \times 1024$ image results in $\hat{cs} = 0.0078$. This leads to the term (1 - $\hat{cs}$) approaching $1$, which diminishes its effectiveness for adaptively re-weighting the loss weights. To address this issue, we design the Boost loss using a scaling factor $\beta$ to balance their distribution for different object sizes.
Our Boost loss can be defined as:
\begin{equation}
\begin{split}
    L_{Boost} = &-\frac{1}{N}\sum_{i=1}^{N}{\{\alpha(1-\hat{cs_i}^{\beta})^\gamma cs_i^{\beta} log(p_{i})} +\\
    &(1-\alpha)p_i^\gamma (1-y_i) log(1 - p_i)\}
\end{split}
  \label{eq:boost}
\end{equation}
where $\alpha$, $\beta$, and $\gamma$ are the hyper-parameters for adjusting the weights of Boost loss, $N$ is the number of objects.

\textbf{Why does the Boost Loss work?} We argue that the key to improving the performance of SOD lies in the model's ability to pay more attention to small objects.
In Boost loss, the smaller size of an object would re-weight the score $p_i$ to a smaller $cs_i$, thereby increasing the loss of small objects and encouraging the model to devote more attention to small objects.
In particular, the weights of object size are exclusively applied to \textit{positive} objects.
The main reason behind this design is that enhancing the loss for negative objects does not strengthen the model's focus on small objects. On the contrary, the number of negative objects is significantly larger than positive objects, which would introduce unnecessary noise into the training process.

\begin{figure*}[t]
  \centering
   \includegraphics[width=0.92\linewidth]{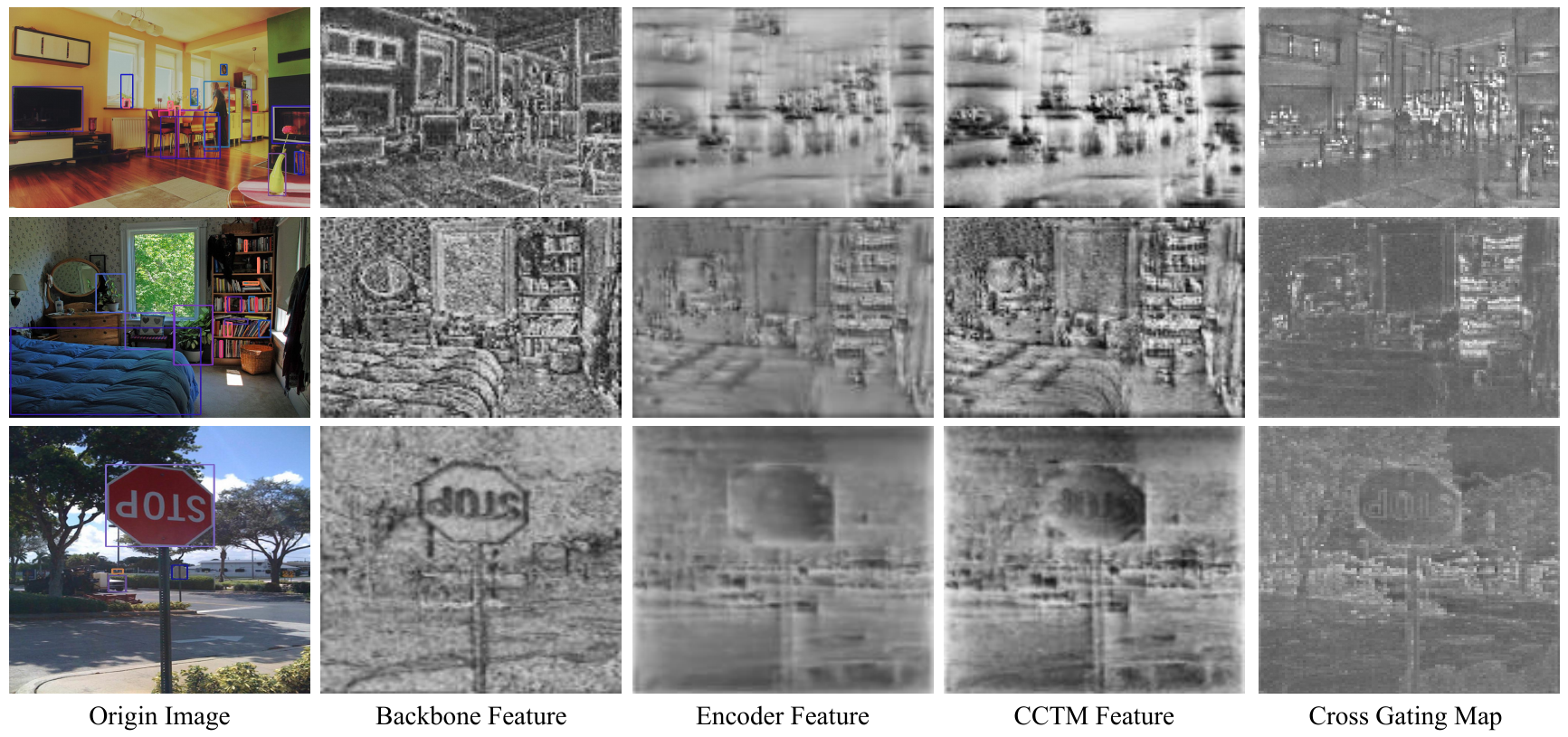}
   \caption{Comparison of the features from different parts of the Cross-DINO model. The visualization shows one channel of these features. The backbone features are extracted using our CLAP-Strip-T model. To enhance the visibility of the details in the Cross Gating Map (column 5), the contrast of the image is adjusted.
   The Cross Gating Map represents the \textit{\textbf{B}}$'$\textit{\textbf{E}}$'$ of CCTM in Fig.~\ref{fig:cctm}. It can be observed that the CCTM feature obtains more details from the backbone feature compared to the original encoder feature.}
   \label{fig:vis}
\end{figure*}

\section{Experiments}\label{sec:experiments}

\begin{table*}[t]

    \caption{Comparison results of Cross-DINO with other popular detection models on \texttt{val2017} of COCO using different backbones and training epochs. Models of DINO and Cross-DINO adopt 4-scales of feature maps from the backbone network.}
\centering
\resizebox{2.0\columnwidth}{!}{
\setlength{\tabcolsep}{1.0mm}{
 \renewcommand\arraystretch{0.98}
\begin{tabular}{ c  c c  c c c c c c c c } 
    \toprule
     Model & Backbone & Epochs & AP & AP$_{50}$ & AP$_{75}$ & AP$_S$ & AP$_{M}$ & AP$_{L}$ & Params & GFLOPs \\
     \midrule
    Faster-RCNN~\cite{ren2015faster} & ResNet50 & 108 & 42.0 & 62.4 & 44.2 & 20.5 & 45.8 & 61.1 & 42M & 180G \\
    DETR~\cite{carion2020end} & ResNet50 & 500 & 42.0 & 62.4 & 44.2 & 20.5 & 45.8 & 61.1 & 41M & 86G \\
    DETR-DC5~\cite{carion2020end} & ResNet50 & 500 & 43.3 & 63.1 & 45.9 & 22.5 & 47.3 & 61.1 & 41M & 187G \\
    Deformable-DETR~\cite{zhu2020deformable} & ResNet50 & 50 & 46.2 & 65.2 & 50.0 & 28.8 & 49.2 & 61.7 & 40M & 173G \\
    Efficient-DETR~\cite{yao2021efficient} & ResNet50 & 36 & 44.2 & 62.2 & 48.0 & 28.4 & 47.5 & 56.6 & 32M & 159G \\
    Conditional DETR~\cite{meng2021conditional} & ResNet50 & 108 & 43.0 & 64.0 & 45.7 & 22.7 & 46.7 & 61.5 & 44M & 90G \\

    Sparse-DETR~\cite{roh2021sparse} & ResNet50 & 50 & 46.3 & 66.0 & 50.1 & 29.0 & 49.5 & 60.8 & 41M & 136G \\
    DAB-DETR~\cite{liu2021dab} & ResNet50 & 50 & 42.6 & 63.2 & 45.6 & 21.8 & 46.2 & 61.1 & 44M & 100G \\
    DN-DETR~\cite{li2022dn} & ResNet50 & 50 & 44.1 & 64.4 & 46.7 & 22.9 & 48.0 & 63.4 & 44M & 94G \\
    CF-DETR~\cite{cao2022cf} & ResNet50 & 36 & 47.8 & 66.5 & 52.4 & 31.2 & 50.6 & 62.8 & - & -\\
    DiffusionDet~\cite{chen2023diffusiondet} & ResNet50 & 60 & 46.8 & 65.3 & 51.8 & 29.6 & 49.3 & 62.2 & - & -\\
    Focus-DETR~\cite{zheng2023less} & ResNet50 & 36 & 50.4 & 68.5 & 55.0 & 34.0 & 53.5 & 64.4 & 48M & 154G \\

    \midrule
    DETR~\cite{carion2020end} & ResNet101 & 500 & 43.5 & 63.8 & 46.4 & 21.9 & 48.0 & 61.8 & 60M & 152G \\    
    Conditional DETR~\cite{meng2021conditional} & ResNet101 & 108 & 44.5 & 65.6 & 47.5 & 23.6 & 48.4 & 63.6 & 63M & 156G \\
    Efficient-DETR~\cite{yao2021efficient} & ResNet101 & 36 & 45.2 & 63.7 & 48.8 & 28.8 & 49.1 & 59.0 & 51M & 239G \\
    DAB-DETR~\cite{liu2021dab} & ResNet101 & 50 & 44.1 & 64.7 & 47.2 & 24.1 & 48.2 & 62.9 & 63M & 179G \\
    DN-DETR~\cite{li2022dn} & ResNet101 & 50 & 45.2 & 65.5 & 48.3 & 24.1 & 49.1 & 65.1 & 63M & 174G \\
    CF-DETR~\cite{cao2022cf} & ResNet101 & 36 & 49.0 & 68.1 & 53.4 & 31.4 & 52.2 & 64.3 & - & - \\
    DiffusionDet~\cite{chen2023diffusiondet} & ResNet101 & 60 & 47.5 & 65.7 & 52.0 & 30.8 & 50.4 & 63.1 & - & - \\

    \midrule
    DINO~\cite{zhang2022dino} & ResNet50 & 12 & 49.0 & 66.6 & 53.5 & 32.0 & 52.3 & 63.0 & 47M & 279G  \\
    Grounding DINO~\cite{liu2023grounding} & ResNet50 & 12 & 48.1 & 65.8 & 52.3 & 30.4 & 51.3 & 62.3 & - & - \\
    Co-Deformable-DETR~\cite{zong2023detrs} & ResNet50 & 12 & 49.5 & 67.6 & 54.3 & 32.4 & 52.7 & 63.7 & - & - \\
    \textbf{Cross-DINO (ours)} & ResNet50 & 12 & \textbf{50.1 (+1.1)} & \textbf{68.0} & \textbf{54.6} & \textbf{33.1 (+1.1)} & \textbf{53.4} & \textbf{65.2} & 48M & 288G\\
    \midrule

    DINO~\cite{zhang2022dino} & ResNet50 & 24 & 50.4 & 68.3 & 54.8 & 33.3 & 53.7 & 64.8 & 47M & 279G  \\
    DINO~\cite{zhang2022dino} & ResNet50 & 36 & 50.9 & 69.0 & 55.3 & 34.6 & 54.1 & 64.6 & 47M & 279G \\
    \textbf{Cross-DINO (ours)} & ResNet50 & 24 & \textbf{51.4 (+1.0)} & \textbf{69.3} & \textbf{56.0} & \textbf{34.1 (+0.8)} & \textbf{54.9} & \textbf{65.5} & 48M & 288G\\
    \midrule

    DINO~\cite{ren2023detrex} & Swin-T & 12 & 51.3 & 69.0 & 56.0 & 34.5 & 54.4 & 66.0 & 48M & 280G  \\
    $\mathcal{H}$-Deformable-DETR~\cite{jia2023detrs} & Swin-T & 12 & 50.6 & 68.9 & 55.1 & 33.4 & 53.7 & 65.9 & - & - \\
    
    \textbf{Cross-DINO (ours)} & Swin-T & 12 & \textbf{52.1 (+0.8)} & \textbf{70.3} & \textbf{56.7} & \textbf{36.9 (+2.4)} & \textbf{55.4} & \textbf{67.0} & 49M & 302G  \\
    \midrule

    $\mathcal{H}$-Deformable-DETR~\cite{jia2023detrs} & Swin-T & 36 & 53.2 & 71.5 & 58.2 & 35.9 & 56.4 & 68.2 & - & - \\
    \textbf{Cross-DINO (ours)} & Swin-T & 36 & \textbf{54.6 (+1.4)} & \textbf{72.7} & \textbf{59.6} & \textbf{37.5 (+1.6)} & \textbf{58.3} & \textbf{69.5} & 49M & 302G \\
    \midrule

    DINO~\cite{zhang2022dino} & ResNet50 & 12 & 49.0 & 66.6 & 53.5 & 32.0 & 52.3 & 63.0 & 47M & 279G  \\
    \textbf{DINO-CLAP-Strip (ours)} & \textbf{CLAP-Strip-T} & 12 & 51.7 (+1.7) & 69.6 & 56.8 & 35.0 (+3.0)& 55.1 & 66.0 & 44M & 263G \\
    \textbf{Cross-DINO (ours)} & \textbf{CLAP-Strip-T} & 12 & 52.6 (+3.6) & 70.7 & 57.6 & 36.4 (+4.4) & 56.6 & 68.2 & 45M & 277G \\

    \textbf{Cross-DINO (ours)} & \textbf{CLAP-Strip-T} & 36 & \textbf{54.6} & \textbf{72.8} & \textbf{59.7} & \textbf{37.5} & \textbf{58.4} & \textbf{69.4} & 45M & 277G  \\
    \bottomrule
     \end{tabular}}
     }
 
\label{tab:sota}
\end{table*}

\subsection{Experiment Setup}

\paragraph{Datasets.}
We evaluate our method on \textit{five} object detection datasets: COCO2017~\cite{lin2014microsoft}, WiderPerson~\cite{zhang2019widerperson}, VisDrone2019~\cite{zhu2021detection}, AI-TOD~\cite{wang2021tiny}, and SODA-D~\cite{cheng2023towards}. These datasets differ significantly in terms of the number of training images, object sizes, and the variety of detection scenes. The ablation studies are conducted on both the COCO2017 and VisDrone2019 dataset.

\textbf{COCO2017.} The COCO~\cite{lin2014microsoft} dataset is a widely used benchmark dataset for object detection. 
COCO2017 consists of 118k training images and 5k validation images, with more than 80 object categories.

\textbf{WiderPerson.} WiderPerson~\cite{zhang2019widerperson} is a large and diverse dataset for dense pedestrian detection in real-world settings. It consists of 13,382 images with a total number of 399,786 annotations, averaging 29.87 annotations per image. This dataset presents significant challenges for SOD due to its diverse scenarios and substantial occlusion. It includes 8,000 images for training and 1,000 images for validation.

\textbf{VisDrone2019.} The VisDrone2019~\cite{zhu2021detection} dataset comprises a total of 10209 images, divided into three subsets: 6,471 images for the training subset, 548 images for the validation subset, and 1,610 for the test-dev subset. This dataset encompasses 10 categories. In our study, we adhere to the training protocols outlined in the SR-TOD~\cite{cao2025visible} and RFLA~\cite{xu2022rfla} methods. We utilize the \textit{training subset} for model training and reserve the \textit{test-dev subset} for evaluating performance.

\textbf{AI-TOD.} The AI-TOD~\cite{wang2021tiny} dataset contains 28,036 aerial images, featuring a total of 700,621 object instances across 8 categories. Each image in the dataset has a resolution of 800 $\times$ 800 pixels. This dataset is particularly challenging due to the small object sizes typical in aerial images, with an average object size of 12.8 pixels.

\textbf{SODA-D.} The SODA-D~\cite{cheng2023towards} dataset includes 24,828 high-quality traffic images and 278,433 instances across 9 categories. Due to the high-resolution (approximately 4000 $\times$ 3000 pixels) of the images in SODA, we applied the pre-processing method from the previous work~\cite{yuan2023small_cfi} to this dataset. Specifically, we divide the original images into 800 $\times$ 800 patches with a stride of 650 pixels and then resize these patches to 1200 $\times$ 1200 pixels for both training and testing. Following the approach in~\cite{yuan2023small_cfi}, we utilize the \textit{training set} for model training and the \textit{test set} for evaluation.

\paragraph{Implementation Details.} 
To ensure a fair comparison, we adopt the same training recipe used in various studies: DINO~\cite{zhang2022dino} for COCO, IterDet~\cite{rukhovich2021iterdet} for WiderPerson, and both SR-TOD~\cite{cao2025visible} and RFLA~\cite{xu2022rfla} for the VisDrone2019 and AI-TOD datasets. Addicionally, we follow the training recipe from CFINet~\cite{yuan2023small_cfi} for the SODA-D dataset. All models are trained using the AdamW~\cite{loshchilov2017decoupled} optimizer with a weight decay of 1 $\times 10^{-4}$. Cross-DINO utilizes \textit{4-scale} features from the backbone. Similar to other detection models~\cite{he2017mask, tang2022image, zhang2022dino}, all backbones in this study are pre-trained on ImageNet-1K~\cite{deng2009imagenet}.
For COCO and WiderPerson, the models are trained with a mini-batch size of 8 on Tesla V100 GPUs. 
By default, unless otherwise specified, $\beta$ is set to 1.0, $\alpha$ is set to 0.25, and $\gamma$ is set to 2.0.
For the SODA-D and VisDrone2019 datasets, the batch size is set to 4. In the case of the AI-TOD dataset, the batch size is set to 2, following the setting outlined in SR-TOD~\cite{cao2025visible} and RFLA~\cite{xu2022rfla}.
The number of decoder queries and denoising queries are set to 900 and 100, respectively, following the DINO approach. In the ablation study, our models are trained for 12 epochs (1$\times$ training scheduler) unless otherwise specified.

\paragraph{Evaluation Criteria.} \textbf{For COCO2017}, we evaluate the detection performance using the standard average precision (AP)~\cite{liu2021survey} metric under various IoU thresholds and object scales.
\textbf{For WiderPerson}, we employ the evaluation metrics of AP, Recall, and mMR, which are commonly used in pedestrian detection~\cite{zhang2019widerperson}. 
The mMR metric, introduced in~\cite{dollar2011pedestrian}, represents the log-average miss rate over false positives per image, ranging from $10^{-2}$ to $10^0$. It serves as a valuable indicator for real-world applications.

For the AI-TOD dataset, object sizes are categorized as follows: sizes ranging from 2 to 8 pixels are classified \textit{very tiny}, 8 to 16 pixels as \textit{tiny}, 16 to 32 pixels as \textit{small}.
Following the methodologies outlined in previous works~\cite{wang2021tiny,cao2025visible,xu2022rfla}, we evaluate both the \textbf{VisDrone2019 and AI-TOD} datasets by using the metrics AP$_{vt}$, AP$_t$, and AP$_s$ to assess performance for the \textit{very tiny}, \textit{tiny}, and \textit{small} object categories, respectively.
For the \textbf{SODA-D} dataset, we adopt the evaluation metrics from previous works~\cite{cheng2023towards,yuan2023small_cfi}, employing AP$_{eS}$ for \textit{extremely Small} objects, AP$_{rS}$ for \textit{relatively Small} objects, and AP$_{gS}$ \textit{for generally Small} objects, where these metrics are specifically introduced for small objects.

\subsection{Main Results}

\paragraph{Results on COCO2017} 
Table.~\ref{tab:sota} compares Cross-DINO on \texttt{val2017} of COCO with different popular detection frameworks, including CNN-based detectors and DETR-like detectors.
It can be observed that Cross-DINO achieves higher performance across all metrics under three \textit{different} backbone models, such as ResNet50, Swin-T and CLAP-Strip-T.
Specifically, when trained with the same settings, such as 12 epochs and the ResNet50 backbone model, Cross-DINO achieves higher performance with \textbf{+1.1\%} increase in AP and \textbf{+1.1\%} increase in AP$_S$ compared to original DINO model.
We also construct a new model called DINO-CLAP-Strip-T by solely replacing the Resnet50 backbone of DINO with introduced CLAP-Strip-T model.
DINO-CLAP-Strip-T achieves higher performance by \textbf{+2.7\%} AP (51.7\% vs. 49.0\%) than DINO, providing strong evidence of the effectiveness of our method.
In addition, building upon the new DINO-CLAP-Strip-T baseline, the inclusion of CCTM and Boost loss in the Cross-DINO model obtains the higher performance of \textbf{52.6\%} AP compared to original DINO model, with an further improvement by \textbf{+0.9\%} AP (totally in \textbf{+3.6\%} AP than DINO).

For SOD, Cross-DINO with ResNet50 and Swin-T models consistently outperforms other advanced models by an average increasing of \textbf{+0.98\%} in AP$_S$.
Furthermore, Cross-DINO with CLAP-Strip-T backbone achieves the performance of \textbf{36.4\%} AP$_S$, with an increase of \textbf{+4.4\%} AP$_S$ than DINO, surpassing other advanced models.

\paragraph{Results on WiderPerson} 
We refer to the previous works~\cite{zhang2019widerperson, rukhovich2021iterdet} for results obtained on the `Hard' subset of annotations, which includes all the boxes with a physical height over 20 pixels.
Table~\ref{tab:widerperson} compares our method with both CNN-based and transformer-based detectors. When using different backbone networks, 
including ResNet50~\cite{he2016deep}, Swin-T~\cite{liu2021swin}, and CLAP-Strip-T, 
our models achieve the higher performance on all evaluation metrics compared to most advanced methods. 
In particular, our Cross-DINO-Swin-T model achieves an AP of \textbf{93.93\%} and a Recall of \textbf{99.65\%}, which is \textit{higher} than the original DINO-Swin-T model by \textbf{+0.86\%} in AP and \textbf{+0.23\%} in Recall, respectively. In addition, our Cross-DINO-CLAP-Strip-T obtains \textbf{36.91\%} in mMR, outperforming the original DINO by \textbf{3.17\%} in mMR.
These results clearly demonstrate the effectiveness of our CLAP-Strip-MLP model, CCTM and Boost loss in enhancing SOD.

\begin{table}[t]
\caption{Experimental results on WiderPerson dataset. 
}
\centering
\resizebox{1.0\columnwidth}{!}{
\setlength{\tabcolsep}{0.5mm}{
 \renewcommand\arraystretch{1.0}
\begin{tabular}{ c c c c c }
    \toprule
     Method & Epochs & AP$\uparrow$ & Recall$\uparrow$ & mMR$\downarrow$ \\
     \midrule
    PS-RCNN~\cite{ge2020ps} & 12 & 89.96 & 94.71 & - \\
    IterDet-1-iter~\cite{rukhovich2021iterdet} & 24 & 89.49 & 92.67 & 40.35 \\ 
    IterDet-2-iter~\cite{rukhovich2021iterdet} & 24 & 91.95 & 97.15 & 40.78 \\
    He et al.~\cite{he2022multi} & - & 91.29 & - & 40.43 \\
    Cascade Transformer~\cite{ma2023cascade} & 50 &  92.98 & 97.66 & 38.41 \\
    \midrule
    DINO-ResNet50~\cite{zhang2022dino} & 24 & 92.75 & 99.08 & 40.08 \\
    \textbf{Cross-DINO-ResNet50 (ours)} & 24 & \textbf{93.29 (+0.54)} & \textbf{99.64} & \textbf{39.94} \\
    \midrule
    DINO-Swin-T~\cite{zhang2022dino} & 24 & 93.07 & 99.42 & 38.78 \\
    \textbf{Cross-DINO-Swin-T (ours)} & 24 & \textbf{93.93 (+0.86)} & \textbf{99.65} & \textbf{37.88} \\
    \midrule
    DINO-CLAP-Strip-T (ours) & 24 & 93.19 & 99.42 & 38.21 \\
    \textbf{Cross-DINO-CLAP-Strip-T (ours)} & 24 & \textbf{93.92 (+0.73)} & \textbf{99.65} & \textbf{36.91} \\
     \bottomrule
     \end{tabular}}
     }
\label{tab:widerperson}
\end{table}

\begin{figure}[t]
  \centering
   \includegraphics[width=0.8\linewidth]{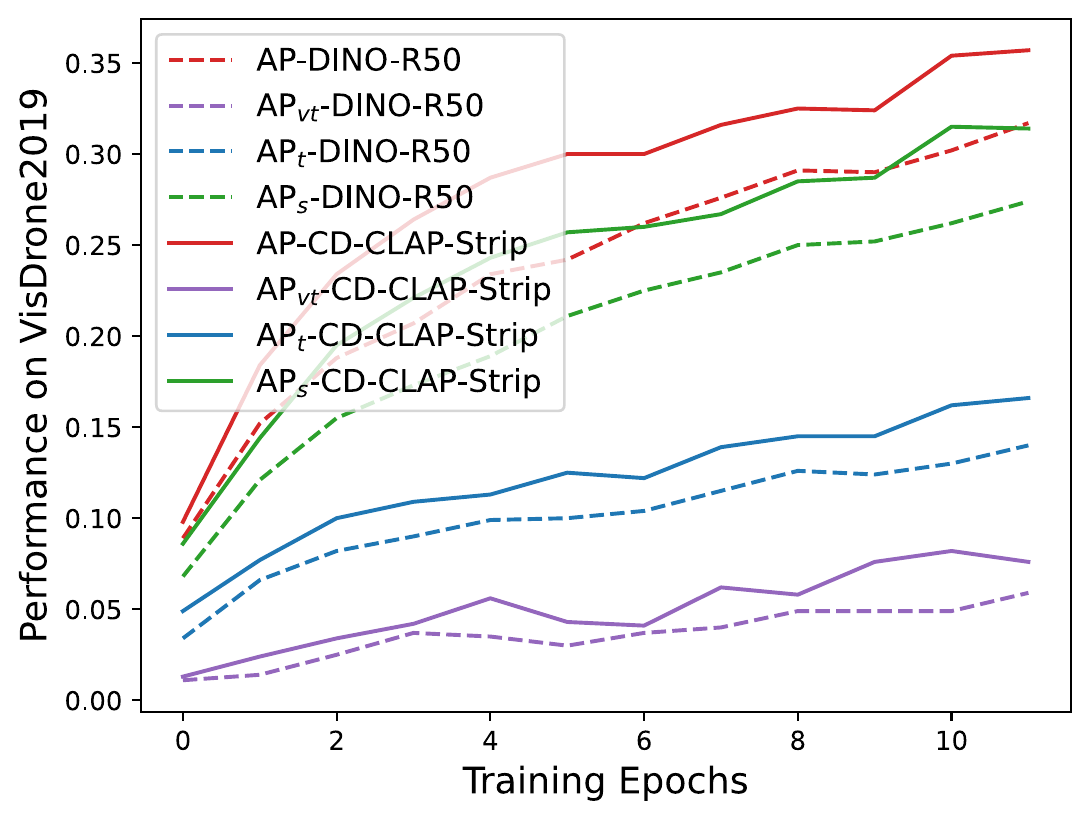}
   \caption{The performance curves for DINO-R50 and Cross-DINO (abbreviated as CD-CLAP-Strip in the legend) on the VisDrone2019 test set.}
   \label{fig:visdrone_training_acwo}
\end{figure}

\paragraph{Results on VisDrone2019} The comparison results for VisDrone2019 dataset are presented in Table~\ref{tab:visdrone}. Compared to the baseline DINO~\cite{zhang2022dino} model, using the ResNet50 backbone, our model improves the overall performance, achieving increases of \textbf{+1.4\%} in AP (33.1\% vs. 31.7\%). Additionally, by incorporating our CLAP-Strip-T model, the DINO model further enhances its accuracy, resulting in increases of \textbf{+1.9\%} in AP and \textbf{+0.6\%} in AP$_{vt}$. Overall, compared to the DINO-R50 baseline model, our Cross-DINO-CLAP-Strip-T model achieves significant improvements of \textbf{+3.7\%} in AP, \textbf{+2.3\%} in AP$_{vt}$, and \textbf{+4.1\%} in AP$_{s}$. Furthermore, as illustrated in Fig.~\ref{fig:visdrone_training_acwo}, our Cross-DINO model (in \textit{solid} line) outperformed the DINO model (in \textit{dashed} line) by a substantial margin during training process, clearly demonstrating its effectiveness.

\begin{table}[h]
\small
\centering
\caption{Comparison results on VisDrone2019 test dataset. }
\resizebox{1.0\columnwidth}{!}{
\setlength{\tabcolsep}{1.0mm}{
 \renewcommand\arraystretch{1.0}
\begin{tabular}{ c | c | c c c c c c }
    \toprule
    Model & Eps & AP & AP$_{0.5}$ & AP$_{0.75}$ & AP$_{vt}$ & AP$_{t}$ & AP$_{s}$ \\
    \midrule
    Faster R-CNN~\cite{ren2015faster} & 12 & 23.9 & 42.2 & 23.8 & 0.1 & 6.5 & 21.1 \\
    Cascade R-CNN~\cite{cai2018cascade} & 12 & 25.2 & 42.6 & 25.9 & 0.1 & 7.0 & 22.5 \\
    DetectoRS~\cite{qiao2021detectors} & 12 & 26.3 & 43.9 & 26.9 & 0.1 & 7.5 & 23.3 \\
    
    \midrule
    Faster R-CNN w/SR-TOD~\cite{cao2025visible} & 12 & 26.3 & 46.8 & 26.0 & 2.9 & 11.0 & 23.7 \\
    DetectoRS w/ SR-TOD~\cite{cao2025visible} & 12 & 27.2 & 47.1 & 27.2 & 2.4 & 11.7 & 24.2 \\
    Cascade R-CNN w/ SR-TOD~\cite{cao2025visible} & 12 & 27.3 & 46.9 & 27.5 & 2.3 & 11.5 & 24.7 \\
    RFLA~\cite{xu2022rfla} & 12 & 27.2 & 48.0 & 26.6 & 4.5 & 13.0 & 23.6 \\
    RFLA w/ SR-TOD~\cite{cao2025visible} & 12 & 27.8 & 48.8 & 27.5 & 4.8 & 13.2 & 24.5 \\
    \midrule
    
    DINO-R50~\cite{zhang2022dino} & 12 & 31.7 & 54.1 & 31.9 & \textbf{5.9} & 14.0 & 27.4 \\ %
   
    \textbf{Cross-DINO-R50} & 12 & \textbf{33.1} & \textbf{55.8} & \textbf{33.7} & 5.7 & \textbf{14.5} & \textbf{28.5} \\
   
    \midrule
    
    \textbf{DINO-CLAP-Strip-T} & 12 & 33.6 & 56.1 & 34.2 & 6.5 & 16.1 & 29.8 \\
    
    \textbf{Cross-DINO-CLAP-Strip-T} & 12 & \textbf{35.4} & \textbf{59.8} & \textbf{35.9} & \textbf{8.2} & \textbf{16.2} & \textbf{31.5} \\
    \midrule
    $\triangle$ Improvement & - & \textcolor{black}{3.7} & \textcolor{black}{5.7} & \textcolor{black}{4.0} & \textcolor{black}{2.3} & \textcolor{black}{2.2} & \textcolor{black}{4.1} \\
     \bottomrule
     \end{tabular}}
     }
\label{tab:visdrone}
\end{table}

\paragraph{Results on AI-TOD} The experimental results in Table~\ref{tab:AI_TOD} indicate that Cross-DINO-R50 achieved a slight overall improvement of \textbf{+0.8\%} in AP. 
Compared to other datasets like VisDrone2019, the improvement on AI-TOD is relatively small. This can be attributed to two main factors: (1) AI-TOD contains a mean object size of 12.8 pixels, which makes SOD more challenging, and (2) the Boost loss may be less effective because these smaller object sizes in AI-TOD lead to similar weights, diminishing its ability to differentiate and prioritize objects effectively.
When employing the proposed CLAP-Strip-T model, the performance drops to 22.6\% AP. We attribute this decline primarily to the \textit{significantly fewer feature channels} in the CLAP-Strip-T compared to ResNet50. Specifically, ResNet50 generates features channels of $[512, 1024, 2048]$, providing more necessary information for detecting tiny objects. In contrast, the CLAP-Strip-T generates features with only $[160, 320, 640]$ channels, resulting in insufficient information for detecting \textit{tiny} and \textit{very tiny} objects.

This issue arises particularly in datasets where the objects are very small, such as the AI-TOD~\cite{wang2021tiny} dataset, which has a mean object size of only 12.8 pixels. This dataset contains a substantial proportion—\textbf{85.6\%}—of \textit{very tiny} (ranging from 2 to 8 pixels) and \textit{tiny} (ranging from 8 to 16 pixels) objects. The small size of these objects presents a significant challenge for the backbone to extract necessary information for effective detection. As a comparison, we resized the input images from 800 $\times$ 800 to 1200 $\times$ 1200, following a similar approach to that used in SODA-D~\cite{cheng2023towards}. As shown in Table~\ref{tab:AI_TOD}, the Cross-DINO-CLAP-Strip-T-1200 model achieved an AP of 25.1\%, with accuracy improvements of +4.5\% for \textit{very tiny} objects and \textit{+2.2\%} for \textit{tiny} objects. 
This result highlights the importance of aggregating information from small objects to enhance the detection accuracy. Additionally, it also suggests that the reduction in accuracy observed with the CLAP-Strip-T model may be primarily due to its fewer feature channels (31.25\%) compared to ResNet50.

Notably, in other datasets such as COCO2017, WiderPerson, VisDrone2019, and SODA-D (in Table~\ref{tab:SODA_D}), our Cross-DINO with CLAP-Strip-T model achieved significant improvements in performance: +4.4\% AP$_S$ in COCO, +1.17\% AP in WiderPerson, +4.3\% AP in SODA-D, and +3.7\% AP in VisDrone2019.
These enhancements can be partially attributed to the effective aggregation of both global and local information (as illustrated in Fig.~\ref{fig:erf_strip_mlp}). Notably, this was achieved while utilizing only 31.25\% of the feature channels compared to ResNet50 (e.g., [160, 320, 640] compared to [512, 1024, 2048]). This demonstrates that the CLAP-Strip-T model significantly enhances the initial feature representation for detectors, highlighting its effectiveness in improving detection performance.

\begin{table}[h]
\small
\centering
\caption{Comparison results on AI-TOD test dataset. The input size for Cross-DINO-CLAP-Strip-T-1200 is 1200 $\times$ 1200, while other models use an input size of 800 $\times$ 800. The $\beta$ is set to 0.05.}
\resizebox{1.0\columnwidth}{!}{
\setlength{\tabcolsep}{0.5mm}{
 \renewcommand\arraystretch{1.0}
\begin{tabular}{ c | c | c c c c c c }
    \toprule
    Model & Eps & AP & AP$_{0.5}$ & AP$_{0.75}$ & AP$_{vt}$ & AP$_{t}$ & AP$_{s}$ \\
    \midrule
    Faster R-CNN~\cite{ren2015faster} & 12 & 11.7 & 27.4 & 8.2 & 0.0 & 8.6 & 23.7 \\
    Cascade R-CNN~\cite{cai2018cascade} & 12 & 14.0 & 31.2 & 10.7 & 0.1 & 10.3 & 26.2 \\
    DetectorRS~\cite{qiao2021detectors} & 12 & 14.6 & 31.8 & 11.5 & 0.0 & 11.0 & 27.4 \\
    RFLA~\cite{xu2022rfla} & 12 & 21.7 & 50.5 & 15.3 & 8.3 & 21.8 & 26.3 \\
    HANet~\cite{guo2023save_hanet} & 12 & 22.1 & 53.7 & 14.4 & 10.9 & 22.2 & 27.3 \\
    RFLA w/ SR-TOD~\cite{cao2025visible} & 12 & 21.8 & 50.8 & 15.4 & 9.7 & 21.8 & 27.4 \\
    Cascade R-CNN w/ SR-TOD~\cite{cao2025visible}& 12 & 21.9 & 50.6 & 15.6 & 9.6 & 22.4 & 26.7 \\
    DetectoRS w/ SR-TOD~\cite{cao2025visible} & 12 & 24.0 & 54.6 & 17.1 & 10.1 & 24.8 & 29.3 \\
    \midrule
    DINO-R50 & 12 & 23.1 & 54.6 & \textbf{16.0} & \textbf{10.5} & 23.5 & 29.1 \\
    \textbf{Cross-DINO-R50} & 12 & \textbf{23.9} & \textbf{57.8} & 15.6 & 10.3 & \textbf{24.5} & \textbf{30.5} \\

    \midrule
    \textbf{Cross-DINO-CLAP-Strip-T} & 12 & 22.6 & 54.7 & 15.4 & 9.5 & 23.5 & 27.7 \\
    \textbf{Cross-DINO-CLAP-Strip-T-1200} & 12 & 25.1 & 58.9 & 17.5 & 13.2 & 25.4 & 31.2 \\
     \bottomrule
     \end{tabular}}
     }
\label{tab:AI_TOD}
\end{table}

\paragraph{Results on SODA-D.} Table~\ref{tab:SODA_D} presents the comparison results between our model and other popular approaches. 
Notably, our Cross-DINO model outperforms other methods, achieving the best performance across all evaluation metrics. Specifically, it surpasses the baseline model DINO~\cite{zhang2022dino} by \textbf{+4.3\%} in AP, \textbf{+3.3\%} in AP$_{eS}$, and \textbf{+5.0\%} in AP$_{N}$. Furthermore, it also outperforms the existing state-of-the-art model CFINet~\cite{yuan2023small_cfi} by \textbf{+1.3\%} in AP and \textbf{+0.3\%} in AP$_{eS}$, and \textbf{+3.3\%} in AP$_N$.
These significant improvements highlight the enhanced capabilities of the Cross-DINO model in transformer-based detectors for the detection of small objects.

\begin{table}[h]
\small
\centering
\caption{Comparison results on the SODA-D test dataset. $\beta$ is set to $0.1$.
}
\resizebox{1.0\columnwidth}{!}{
\setlength{\tabcolsep}{1.0mm}{
 \renewcommand\arraystretch{1.0}
\begin{tabular}{ c | c | c c c | c c c c }
    \toprule
    Model & Eps & AP & AP$_{0.5}$ & AP$_{0.75}$ & AP$_{eS}$ & AP$_{rS}$ & AP$_{gS}$ & AP$_{N}$ \\
    \midrule
    RetinaNet~\cite{lin2017focal} & 12 & 28.2 & 57.6 & 23.7 & 11.9 & 25.2 & 34.1 & 44.2 \\
    DyHead~\cite{dai2021dynamic} & 12 & 27.5 & 56.1 & 23.2 & 12.4 & 24.4 & 33.0 & 41.9 \\
    
    \midrule
    CornerNet~\cite{law2018cornernet} & 24 & 24.6 & 49.5 & 21.7 & 6.5 & 20.5 & 32.2 & 43.8 \\
    RepPoints~\cite{yang2019reppoints} & 12 & 28.0 & 55.6 & 24.7 & 10.1 & 23.8 & 35.1 & 45.3 \\
    \midrule
    Deformable-DETR~\cite{zhu2020deformable} & 50 & 19.2 & 44.8 & 13.7 & 6.3 & 15.4 & 24.9 & 34.2 \\
    Sparse RCNN~\cite{sun2021sparse} & 12 & 24.2 & 50.3 & 20.3 & 8.8 & 20.4 & 30.2 & 39.4 \\
    \midrule
    RFLA~\cite{xu2022rfla} & 12 & 29.7 & 60.2 & 25.2 & 13.2 & 26.9 & 35.4 & 44.6 \\
    CFINet~\cite{yuan2023small_cfi} & 12 & 30.7 & 60.8 & 26.7 & 14.7 & 27.8 & 36.4 & 44.6 \\
    \midrule
    DINO~\cite{zhang2022dino} & 12 & 27.7 & 55.5 & 23.8 & 11.7 & 23.7 & 33.9 & 42.9 \\

    \textbf{Cross-DINO (ours)} & 12 & \textbf{32.0} & \textbf{63.0} & \textbf{27.9} & \textbf{15.0} & \textbf{28.4} & \textbf{38.3} & \textbf{47.9} \\

    \midrule
    $\triangle$ Improvement & - & 4.3 & 7.5 & 4.1 & 3.3 & 4.7 & 4.4 & 5.0 \\
     \bottomrule
     \end{tabular}}
     }
\label{tab:SODA_D}
\end{table}

\subsection{Ablation Study}

\paragraph{Ablation on CCTM and Boost Loss in COCO} 
To further validate the effectiveness of our method, we conduct ablation studies of CCTM and Boost Loss on three different backbone models. 
In Table~\ref{tab:cctm_boost}, comparing the Cross-DINO to the baseline model of DINO using different backbone models, we observe that both CCTM and Boost Loss contribute significantly to improving all evaluation metrics. 
These results highlight the importance of capturing fine-grained details from the backbone features and enhancing the class prediction scores for SOD.
The statistical results in Fig.~\ref{fig:class_pro} also clearly demonstrate that our method with Boost Loss effectively enhances the model's class prediction score than DINO.
When utilizing CLAP-Strip-T as the backbone, applying CCTM and Boost Loss results in a significant improvement of \textbf{+0.7\%} (35.7\% vs. 35.0\%) and \textbf{+1.2\%} (36.2\% vs. 35.0\%) in AP$_S$ for small objects, respectively, compared to DINO-CLAP-Strip-T. Furthermore, by incorporating both CCTM and Boost Loss, we achieve even higher AP and AP$_S$ with an improvement of \textbf{+0.9\%} (52.6\% vs. 51.7\%) and \textbf{+1.4\%} (36.4\% vs. 35.0\%), respectively. The similar improvements are also observed in both the ResNet50 and Swin-T backbone models, highlighting the effectiveness of our method across different architectures.

\begin{table}[t]
\caption{The ablation results of the CCTM and Boost Loss on different backbone networks. For the Swin-T model, the $\beta$ is set to 0.1.}
\centering
\resizebox{1.0\columnwidth}{!}{
\setlength{\tabcolsep}{0.5mm}{
 \renewcommand\arraystretch{1.0}
\begin{tabular}{ c c c c c c c c c c c} \\
    \toprule
     CCTM & Boost & Backbone & AP &  AP$_{50}$ & AP$_{75}$ & AP$_S$ & AP$_{M}$ & AP$_{L}$ \\
     \midrule
    ~ & ~ & ResNet50 & 49.0 & 66.6 & 53.5 & 32.0 & 52.3 & 63.0 \\
    \checkmark & ~ & ResNet50 & 49.8 & 67.6 & 54.5 & 32.4 & 53.0 & 64.3 \\
    ~ & \checkmark & ResNet50 & 49.6 & 67.0 & 54.0 & 32.2 & 53.2 & 64.3 \\
    \checkmark & \checkmark & ResNet50 & 50.0 & 67.8 & 54.7 & 32.6 & 53.5 & 65.4 \\
     \midrule
     ~ & ~ & Swin-T & 51.3 & 69.0 & 56.0 & 34.5 & 54.4 & 66.0 \\
     \checkmark & ~ & Swin-T & 52.0 & 69.8 & 56.9 & 35.4 & 55.1 & 66.8 \\
    ~ & \checkmark & Swin-T & 52.0 & 70.1 & 56.7 & 35.4 & 54.8 & 66.9 \\
    \checkmark & \checkmark & Swin-T & 52.1 & 70.3 & 56.7 & 36.9 & 55.4 & 67.0 \\
     \midrule
     ~ & ~ & CLAP-Strip-T & 51.7 & 69.6 &  56.8 & 35.0 & 55.1 & 66.0 \\
     \checkmark & ~ & CLAP-Strip-T & 52.4 & 70.6 & 57.5 & 35.7 & 56.4 & 67.2 \\
     ~ & \checkmark & CLAP-Strip-T & 52.6 & 70.8 & 57.5  & 36.2 & 56.4 & 68.0 \\
     \checkmark & \checkmark & CLAP-Strip-T & 52.6 & 70.7 & 57.6 & 36.4 & 56.6 & 68.2 \\
     \bottomrule
     \end{tabular}}
     }
\label{tab:cctm_boost}
\end{table}

\paragraph{Ablation on the components of CLAP-Strip-T backbone, CCTM, and Boost Loss in VisDrone2019} To further validate the effectiveness of our method, we conducted additional ablation experiments on the VisDrone2019 dataset. The results presented in Table~\ref{tab:ablation_cross_dino_visdrone} demonstrate that combining the CLAP-Strip-T backbone, CCTM, and Boost Loss methods significantly improves the model's accuracy, with increases of \textbf{+3.7\%} in AP and \textbf{+2.3\%} in AP$_{vt}$ over DINO model. These results further confirm the effectiveness of our approach.

\begin{table}[h]
\caption{The ablation results of the Cross-DINO components. The baseline model used for comparison is DINO-ResNet50.}
\centering
\resizebox{1.0\columnwidth}{!}{
\setlength{\tabcolsep}{0.5mm}{
 \renewcommand\arraystretch{1.0}
\begin{tabular}{ c c c c c c c c c c c} \\
    \toprule
     CLAP-Strip-T & CCTM & Boost & AP &  AP$_{50}$ & AP$_{75}$ & AP$_{vt}$ & AP$_{t}$ & AP$_{s}$ \\
     \midrule
     ~ & ~ & ~ & 31.7 & 54.1 & 31.9 & 5.9 & 14.0 & 27.4 \\ 
     \checkmark & ~ & ~ & 33.6 & 56.1 & 34.2 & 6.5 & 16.1 & 29.8 \\
     \checkmark & \checkmark & ~ & 33.8 & 56.1 & 34.6 & \textbf{8.4} & 15.9 & 29.5 \\
     \checkmark & \checkmark & \checkmark & \textbf{35.4} & \textbf{59.8} & \textbf{35.9}  & 8.2 & \textbf{16.2} & \textbf{31.5} \\
     \bottomrule
     \end{tabular}}
     }
\label{tab:ablation_cross_dino_visdrone}
\end{table}

\paragraph{Ablation on the internal components of CCTM in COCO} The CCTM is designed to adaptively enhance the encoder feature by using the cross-gating operations, selectively and step-by-step fusing the fine-grained information from the backbone features, thereby enriching the details of objects.
To further verify the effectiveness of CCTM, we conduct ablation studies to evaluate the impact of different components in two ways: remove all gating operations and replace them only with MLP layers (MLP-only), and only remove the Gating Map of \textit{\textbf{B}}$'$ and \textit{\textbf{E}}$'$ (once-gating only), respectively. 
Table~\ref{tab:cctm_ablation} clearly indicates that CCTM with twice cross-gating performs the best, \textit{outperforming} once-gating, MLP-only etc.

\begin{table}[h]
\centering
\caption{The ablation results of the CCTM components on COCO.}
\resizebox{1.0\columnwidth}{!}{
\setlength{\tabcolsep}{0.2mm}{
 \renewcommand\arraystretch{1.0}
\begin{tabular}{ c c c c c c c c}
    \toprule
     Model & Ablations & AP & AP$_{50}$ & AP$_{75}$ & AP$_S$ & AP$_{M}$ & AP$_{L}$ \\
    \midrule
    DINO-ResNet50 &  Baseline & 49.0 & 66.6 & 53.5 & 32.0 & 52.3 & 63.0 \\ 
    DINO-ResNet50 & MLP only (w/o gating) & 49.1 & 66.9 & 53.3 & 32.3 & 52.3 & 63.2 \\ 
    DINO-ResNet50 & Once-gating only & 49.4 & 66.8 & 53.9 & 32.1 & 52.9 & 63.5 \\ 
    DINO-ResNet50 & CCTM (twice-gating) & \textbf{49.8} & \textbf{67.6} & \textbf{54.5} & \textbf{32.4} & \textbf{53.0} & \textbf{64.3} \\
     \bottomrule
     \end{tabular}}
     }
\label{tab:cctm_ablation}
\end{table}

\section{Conclusion}\label{sec:conclusion}

This paper proposes a novel Cross-DINO model, addressing the challenges of SOD in DETR-like detection frameworks. We introduce a deep MLP-based model with new CLAP approach into DINO to aggregate both the long-range and short-range feature representations with the context cues for small objects. To alleviate the feature blurring problem caused by multiple attention layers, we present a new CCTM module to feed fine-grained details from backbone features into encoder features. Additionally, we propose a novel efficient Boost loss for enhancing the confidence of class prediction by incorporating the category and size information of objects. Our extensive experimental results demonstrate the effectiveness of Cross-DINO. We hope that our results can spark further research based on the DETR-like framework for OD in the vision community.

{\small
\bibliographystyle{ieee_fullname}
\bibliography{tmm2025}

\begin{thebibliography}{10}\itemsep=-1pt

\bibitem{bosquet2023full}
Brais Bosquet, Daniel Cores, Lorenzo Seidenari, V{\'\i}ctor~M Brea, Manuel Mucientes, and Alberto Del~Bimbo.
\newblock A full data augmentation pipeline for small object detection based on generative adversarial networks.
\newblock {\em Pattern Recognition}, 133:108998, 2023.

\bibitem{cai2018cascade}
Zhaowei Cai and Nuno Vasconcelos.
\newblock Cascade r-cnn: Delving into high quality object detection.
\newblock In {\em Proceedings of the IEEE conference on computer vision and pattern recognition}, pages 6154--6162, 2018.

\bibitem{cao2025visible}
Bing Cao, Haiyu Yao, Pengfei Zhu, and Qinghua Hu.
\newblock Visible and clear: Finding tiny objects in difference map.
\newblock In {\em European Conference on Computer Vision}, pages 1--18. Springer, 2025.

\bibitem{cao2024mlp}
Guiping Cao, Wenjian Huang, Xiangyuan Lan, Jianguo Zhang, Dongmei Jiang, and Yaowei Wang.
\newblock Mlp-dino: Category modeling and query graphing with deep mlp for object detection.
\newblock In {\em IJCAI}, 2024.

\bibitem{cao2023strip}
Guiping Cao, Shengda Luo, Wenjian Huang, Xiangyuan Lan, Dongmei Jiang, Yaowei Wang, and Jianguo Zhang.
\newblock Strip-mlp: Efficient token interaction for vision mlp.
\newblock In {\em Proceedings of the IEEE/CVF International Conference on Computer Vision}, pages 1494--1504, 2023.

\bibitem{cao2022cf}
Xipeng Cao, Peng Yuan, Bailan Feng, and Kun Niu.
\newblock Cf-detr: Coarse-to-fine transformers for end-to-end object detection.
\newblock In {\em Proceedings of the AAAI Conference on Artificial Intelligence}, volume~36, pages 185--193, 2022.

\bibitem{carion2020end}
Nicolas Carion, Francisco Massa, Gabriel Synnaeve, Nicolas Usunier, Alexander Kirillov, and Sergey Zagoruyko.
\newblock End-to-end object detection with transformers.
\newblock In {\em European conference on computer vision}, pages 213--229. Springer, 2020.

\bibitem{chen2021crossvit}
Chun-Fu~Richard Chen, Quanfu Fan, and Rameswar Panda.
\newblock Crossvit: Cross-attention multi-scale vision transformer for image classification.
\newblock In {\em Proceedings of the IEEE/CVF international conference on computer vision}, pages 357--366, 2021.

\bibitem{chen2023diffusiondet}
Shoufa Chen, Peize Sun, Yibing Song, and Ping Luo.
\newblock Diffusiondet: Diffusion model for object detection.
\newblock In {\em Proceedings of the IEEE/CVF International Conference on Computer Vision}, pages 19830--19843, 2023.

\bibitem{cheng2023towards}
Gong Cheng, Xiang Yuan, Xiwen Yao, Kebing Yan, Qinghua Zeng, Xingxing Xie, and Junwei Han.
\newblock Towards large-scale small object detection: Survey and benchmarks.
\newblock {\em IEEE Transactions on Pattern Analysis and Machine Intelligence}, 45(11):13467--13488, 2023.

\bibitem{dai2021dynamic}
Xiyang Dai, Yinpeng Chen, Bin Xiao, Dongdong Chen, Mengchen Liu, Lu Yuan, and Lei Zhang.
\newblock Dynamic head: Unifying object detection heads with attentions.
\newblock In {\em Proceedings of the IEEE/CVF conference on computer vision and pattern recognition}, pages 7373--7382, 2021.

\bibitem{deng2009imagenet}
Jia Deng, Wei Dong, Richard Socher, Li-Jia Li, Kai Li, and Li Fei-Fei.
\newblock Imagenet: A large-scale hierarchical image database.
\newblock In {\em 2009 IEEE conference on computer vision and pattern recognition}, pages 248--255. Ieee, 2009.

\bibitem{dollar2011pedestrian}
Piotr Dollar, Christian Wojek, Bernt Schiele, and Pietro Perona.
\newblock Pedestrian detection: An evaluation of the state of the art.
\newblock {\em IEEE transactions on pattern analysis and machine intelligence}, 34(4):743--761, 2011.

\bibitem{dosovitskiy2020image}
Alexey Dosovitskiy, Lucas Beyer, Alexander Kolesnikov, Dirk Weissenborn, Xiaohua Zhai, Thomas Unterthiner, Mostafa Dehghani, Matthias Minderer, Georg Heigold, Sylvain Gelly, et~al.
\newblock An image is worth 16x16 words: Transformers for image recognition at scale.
\newblock In {\em International Conference on Learning Representations}, 2020.

\bibitem{duan2019centernet}
Kaiwen Duan, Song Bai, Lingxi Xie, Honggang Qi, Qingming Huang, and Qi Tian.
\newblock Centernet: Keypoint triplets for object detection.
\newblock In {\em Proceedings of the IEEE/CVF international conference on computer vision}, pages 6569--6578, 2019.

\bibitem{fu2017dssd}
Cheng-Yang Fu, Wei Liu, Ananth Ranga, Ambrish Tyagi, and Alexander~C Berg.
\newblock Dssd: Deconvolutional single shot detector.
\newblock {\em arXiv preprint arXiv:1701.06659}, 2017.

\bibitem{ge2020ps}
Zheng Ge, Zequn Jie, Xin Huang, Rong Xu, and Osamu Yoshie.
\newblock Ps-rcnn: Detecting secondary human instances in a crowd via primary object suppression.
\newblock In {\em 2020 IEEE international conference on multimedia and expo (ICME)}, pages 1--6. IEEE, 2020.

\bibitem{guo2023save_hanet}
Guangqian Guo, Pengfei Chen, Xuehui Yu, Zhenjun Han, Qixiang Ye, and Shan Gao.
\newblock Save the tiny, save the all: hierarchical activation network for tiny object detection.
\newblock {\em IEEE transactions on circuits and systems for video technology}, 34:221--234, 2023.

\bibitem{he2017mask}
Kaiming He, Georgia Gkioxari, Piotr Doll{\'a}r, and Ross Girshick.
\newblock Mask r-cnn.
\newblock In {\em Proceedings of the IEEE international conference on computer vision}, pages 2961--2969, 2017.

\bibitem{he2016deep}
Kaiming He, Xiangyu Zhang, Shaoqing Ren, and Jian Sun.
\newblock Deep residual learning for image recognition.
\newblock In {\em Proceedings of the IEEE conference on computer vision and pattern recognition}, pages 770--778, 2016.

\bibitem{he2022multi}
Yuzhe He, Ning He, Ren Zhang, Kang Yan, and Haigang Yu.
\newblock Multi-scale feature balance enhancement network for pedestrian detection.
\newblock {\em Multimedia Systems}, 28(3):1135--1145, 2022.

\bibitem{hendrycks2016gaussian}
Dan Hendrycks and Kevin Gimpel.
\newblock Gaussian error linear units (gelus).
\newblock {\em arXiv preprint arXiv:1606.08415}, 2016.

\bibitem{howard2017mobilenets}
Andrew~G Howard, Menglong Zhu, Bo Chen, Dmitry Kalenichenko, Weijun Wang, Tobias Weyand, Marco Andreetto, and Hartwig Adam.
\newblock Mobilenets: Efficient convolutional neural networks for mobile vision applications.
\newblock {\em arXiv preprint arXiv:1704.04861}, 2017.

\bibitem{huang2025_dqdetr}
Yi-Xin Huang, Hou-I Liu, Hong-Han Shuai, and Wen-Huang Cheng.
\newblock Dq-detr: Detr with dynamic query for tiny object detection.
\newblock In {\em European Conference on Computer Vision}, pages 290--305. Springer, 2025.

\bibitem{jia2023detrs}
Ding Jia, Yuhui Yuan, Haodi He, Xiaopei Wu, Haojun Yu, Weihong Lin, Lei Sun, Chao Zhang, and Han Hu.
\newblock Detrs with hybrid matching.
\newblock In {\em Proceedings of the IEEE/CVF Conference on Computer Vision and Pattern Recognition}, pages 19702--19712, 2023.

\bibitem{kisantal2019augmentation}
Mate Kisantal, Zbigniew Wojna, Jakub Murawski, Jacek Naruniec, and Kyunghyun Cho.
\newblock Augmentation for small object detection.
\newblock {\em arXiv preprint arXiv:1902.07296}, 2019.

\bibitem{law2018cornernet}
Hei Law and Jia Deng.
\newblock Cornernet: Detecting objects as paired keypoints.
\newblock In {\em Proceedings of the European conference on computer vision (ECCV)}, pages 734--750, 2018.

\bibitem{li2022dn}
Feng Li, Hao Zhang, Shilong Liu, Jian Guo, Lionel~M Ni, and Lei Zhang.
\newblock Dn-detr: Accelerate detr training by introducing query denoising.
\newblock In {\em Proceedings of the IEEE/CVF Conference on Computer Vision and Pattern Recognition}, pages 13619--13627, 2022.

\bibitem{lin2017focal}
Tsung-Yi Lin, Priya Goyal, Ross Girshick, Kaiming He, and Piotr Doll{\'a}r.
\newblock Focal loss for dense object detection.
\newblock In {\em Proceedings of the IEEE international conference on computer vision}, pages 2980--2988, 2017.

\bibitem{lin2014microsoft}
Tsung-Yi Lin, Michael Maire, Serge Belongie, James Hays, Pietro Perona, Deva Ramanan, Piotr Doll{\'a}r, and C~Lawrence Zitnick.
\newblock Microsoft coco: Common objects in context.
\newblock In {\em Computer Vision--ECCV 2014: 13th European Conference, Zurich, Switzerland, September 6-12, 2014, Proceedings, Part V 13}, pages 740--755. Springer, 2014.

\bibitem{liu2022we}
Ruiyang Liu, Yinghui Li, Linmi Tao, Dun Liang, and Hai-Tao Zheng.
\newblock Are we ready for a new paradigm shift? a survey on visual deep mlp.
\newblock {\em Patterns}, 3(7):100520, 2022.

\bibitem{liu2021dab}
Shilong Liu, Feng Li, Hao Zhang, Xiao Yang, Xianbiao Qi, Hang Su, Jun Zhu, and Lei Zhang.
\newblock Dab-detr: Dynamic anchor boxes are better queries for detr.
\newblock In {\em International Conference on Learning Representations}, 2021.

\bibitem{liu2023detection}
Shilong Liu, Tianhe Ren, Jiayu Chen, Zhaoyang Zeng, Hao Zhang, Feng Li, Hongyang Li, Jun Huang, Hang Su, Jun Zhu, et~al.
\newblock Detection transformer with stable matching.
\newblock In {\em 2023 IEEE/CVF International Conference on Computer Vision (ICCV)}, pages 6468--6477. IEEE, 2023.

\bibitem{liu2023grounding}
Shilong Liu, Zhaoyang Zeng, Tianhe Ren, Feng Li, Hao Zhang, Jie Yang, Chunyuan Li, Jianwei Yang, Hang Su, Jun Zhu, et~al.
\newblock Grounding dino: Marrying dino with grounded pre-training for open-set object detection.
\newblock {\em arXiv preprint arXiv:2303.05499}, 2023.

\bibitem{liu2016ssd}
Wei Liu, Dragomir Anguelov, Dumitru Erhan, Christian Szegedy, Scott Reed, Cheng-Yang Fu, and Alexander~C Berg.
\newblock Ssd: Single shot multibox detector.
\newblock In {\em Computer Vision--ECCV 2016: 14th European Conference, Amsterdam, The Netherlands, October 11--14, 2016, Proceedings, Part I 14}, pages 21--37. Springer, 2016.

\bibitem{liu2021survey}
Yang Liu, Peng Sun, Nickolas Wergeles, and Yi Shang.
\newblock A survey and performance evaluation of deep learning methods for small object detection.
\newblock {\em Expert Systems with Applications}, 172:114602, 2021.

\bibitem{liu2024vmamba}
Yue Liu, Yunjie Tian, Yuzhong Zhao, Hongtian Yu, Lingxi Xie, Yaowei Wang, Qixiang Ye, and Yunfan Liu.
\newblock Vmamba: Visual state space model.
\newblock {\em arXiv preprint arXiv:2401.10166}, 2024.

\bibitem{liu2021swin}
Ze Liu, Yutong Lin, Yue Cao, Han Hu, Yixuan Wei, Zheng Zhang, Stephen Lin, and Baining Guo.
\newblock Swin transformer: Hierarchical vision transformer using shifted windows.
\newblock In {\em Proceedings of the IEEE/CVF international conference on computer vision}, pages 10012--10022, 2021.

\bibitem{liu2022convnet}
Zhuang Liu, Hanzi Mao, Chao-Yuan Wu, Christoph Feichtenhofer, Trevor Darrell, and Saining Xie.
\newblock A convnet for the 2020s.
\newblock In {\em Proceedings of the IEEE/CVF conference on computer vision and pattern recognition}, pages 11976--11986, 2022.

\bibitem{loshchilov2017decoupled}
Ilya Loshchilov and Frank Hutter.
\newblock Decoupled weight decay regularization.
\newblock In {\em International Conference on Learning Representations}, 2018.

\bibitem{luo2016understanding}
Wenjie Luo, Yujia Li, Raquel Urtasun, and Richard Zemel.
\newblock Understanding the effective receptive field in deep convolutional neural networks.
\newblock {\em Advances in neural information processing systems}, 29, 2016.

\bibitem{ma2023cascade}
Chunjie Ma, Li Zhuo, Jiafeng Li, Yutong Zhang, and Jing Zhang.
\newblock Cascade transformer decoder based occluded pedestrian detection with dynamic deformable convolution and gaussian projection channel attention mechanism.
\newblock {\em IEEE Transactions on Multimedia}, 2023.

\bibitem{meng2021conditional}
Depu Meng, Xiaokang Chen, Zejia Fan, Gang Zeng, Houqiang Li, Yuhui Yuan, Lei Sun, and Jingdong Wang.
\newblock Conditional detr for fast training convergence.
\newblock In {\em Proceedings of the IEEE/CVF International Conference on Computer Vision}, pages 3651--3660, 2021.

\bibitem{pang2019efficient}
Yanwei Pang, Tiancai Wang, Rao~Muhammad Anwer, Fahad~Shahbaz Khan, and Ling Shao.
\newblock Efficient featurized image pyramid network for single shot detector.
\newblock In {\em Proceedings of the IEEE/CVF conference on computer vision and pattern recognition}, pages 7336--7344, 2019.

\bibitem{qiao2021detectors}
Siyuan Qiao, Liang-Chieh Chen, and Alan Yuille.
\newblock Detectors: Detecting objects with recursive feature pyramid and switchable atrous convolution.
\newblock In {\em Proceedings of the IEEE/CVF conference on computer vision and pattern recognition}, pages 10213--10224, 2021.

\bibitem{redmon2016you}
Joseph Redmon, Santosh Divvala, Ross Girshick, and Ali Farhadi.
\newblock You only look once: Unified, real-time object detection.
\newblock In {\em Proceedings of the IEEE conference on computer vision and pattern recognition}, pages 779--788, 2016.

\bibitem{rekavandi2023transformers}
Aref~Miri Rekavandi, Shima Rashidi, Farid Boussaid, Stephen Hoefs, Emre Akbas, et~al.
\newblock Transformers in small object detection: A benchmark and survey of state-of-the-art.
\newblock {\em arXiv preprint arXiv:2309.04902}, 2023.

\bibitem{ren2015faster}
Shaoqing Ren, Kaiming He, Ross Girshick, and Jian Sun.
\newblock Faster r-cnn: Towards real-time object detection with region proposal networks.
\newblock {\em Advances in neural information processing systems}, 28, 2015.

\bibitem{ren2023detrex}
Tianhe Ren, Shilong Liu, Feng Li, Hao Zhang, Ailing Zeng, Jie Yang, Xingyu Liao, Ding Jia, Hongyang Li, He Cao, et~al.
\newblock detrex: Benchmarking detection transformers.
\newblock {\em arXiv preprint arXiv:2306.07265}, 2023.

\bibitem{roh2021sparse}
Byungseok Roh, JaeWoong Shin, Wuhyun Shin, and Saehoon Kim.
\newblock Sparse detr: Efficient end-to-end object detection with learnable sparsity.
\newblock In {\em International Conference on Learning Representations}, 2021.

\bibitem{rukhovich2021iterdet}
Danila Rukhovich, Konstantin Sofiiuk, Danil Galeev, Olga Barinova, and Anton Konushin.
\newblock Iterdet: iterative scheme for object detection in crowded environments.
\newblock In {\em Structural, Syntactic, and Statistical Pattern Recognition: Joint IAPR International Workshops, S+ SSPR 2020, Padua, Italy, January 21--22, 2021, Proceedings}, pages 344--354. Springer, 2021.

\bibitem{shou2022object}
Yuntao Shou, Tao Meng, Wei Ai, Canhao Xie, Haiyan Liu, and Yina Wang.
\newblock Object detection in medical images based on hierarchical transformer and mask mechanism.
\newblock {\em Computational Intelligence and Neuroscience}, 2022, 2022.

\bibitem{singh2018sniper}
Bharat Singh, Mahyar Najibi, and Larry~S Davis.
\newblock Sniper: Efficient multi-scale training.
\newblock In {\em Advances in Neural Information Processing Systems}, volume~31. Curran Associates, Inc., 2018.

\bibitem{sun2021sparse}
Peize Sun, Rufeng Zhang, Yi Jiang, Tao Kong, Chenfeng Xu, Wei Zhan, Masayoshi Tomizuka, Lei Li, Zehuan Yuan, Changhu Wang, et~al.
\newblock Sparse r-cnn: End-to-end object detection with learnable proposals.
\newblock In {\em Proceedings of the IEEE/CVF conference on computer vision and pattern recognition}, pages 14454--14463, 2021.

\bibitem{tang2022image}
Yehui Tang, Kai Han, Jianyuan Guo, Chang Xu, Yanxi Li, Chao Xu, and Yunhe Wang.
\newblock An image patch is a wave: Phase-aware vision mlp.
\newblock In {\em Proceedings of the IEEE/CVF Conference on Computer Vision and Pattern Recognition}, pages 10935--10944, 2022.

\bibitem{tian2024multi}
Haoyu Tian, Yipeng Zhang, Hanbo Wu, Xin Ma, and Yibin Li.
\newblock Multi-scale sampling attention graph convolutional networks for skeleton-based action recognition.
\newblock {\em Neurocomputing}, page 128086, 2024.

\bibitem{touvron2021training}
Hugo Touvron, Matthieu Cord, Matthijs Douze, Francisco Massa, Alexandre Sablayrolles, and Herv{\'e} J{\'e}gou.
\newblock Training data-efficient image transformers \& distillation through attention.
\newblock In {\em International conference on machine learning}, pages 10347--10357. PMLR, 2021.

\bibitem{wang2023yolov7}
Chien-Yao Wang, Alexey Bochkovskiy, and Hong-Yuan~Mark Liao.
\newblock Yolov7: Trainable bag-of-freebies sets new state-of-the-art for real-time object detectors.
\newblock In {\em Proceedings of the IEEE/CVF Conference on Computer Vision and Pattern Recognition}, pages 7464--7475, 2023.

\bibitem{wang2021normalized}
Jinwang Wang, Chang Xu, Wen Yang, and Lei Yu.
\newblock A normalized gaussian wasserstein distance for tiny object detection.
\newblock {\em arXiv preprint arXiv:2110.13389}, 2021.

\bibitem{wang2021tiny}
Jinwang Wang, Wen Yang, Haowen Guo, Ruixiang Zhang, and Gui-Song Xia.
\newblock Tiny object detection in aerial images.
\newblock In {\em 2020 25th international conference on pattern recognition (ICPR)}, pages 3791--3798. IEEE, 2021.

\bibitem{woo2023convnext}
Sanghyun Woo, Shoubhik Debnath, Ronghang Hu, Xinlei Chen, Zhuang Liu, In~So Kweon, and Saining Xie.
\newblock Convnext v2: Co-designing and scaling convnets with masked autoencoders.
\newblock In {\em Proceedings of the IEEE/CVF Conference on Computer Vision and Pattern Recognition}, pages 16133--16142, 2023.

\bibitem{xu2022rfla}
Chang Xu, Jinwang Wang, Wen Yang, Huai Yu, Lei Yu, and Gui-Song Xia.
\newblock Rfla: Gaussian receptive field based label assignment for tiny object detection.
\newblock In {\em European conference on computer vision}, pages 526--543. Springer, 2022.

\bibitem{yang2022querydet}
Chenhongyi Yang, Zehao Huang, and Naiyan Wang.
\newblock Querydet: Cascaded sparse query for accelerating high-resolution small object detection.
\newblock In {\em Proceedings of the IEEE/CVF Conference on computer vision and pattern recognition}, pages 13668--13677, 2022.

\bibitem{yang2019reppoints}
Ze Yang, Shaohui Liu, Han Hu, Liwei Wang, and Stephen Lin.
\newblock Reppoints: Point set representation for object detection.
\newblock In {\em Proceedings of the IEEE/CVF international conference on computer vision}, pages 9657--9666, 2019.

\bibitem{yao2021efficient}
Zhuyu Yao, Jiangbo Ai, Boxun Li, and Chi Zhang.
\newblock Efficient detr: improving end-to-end object detector with dense prior.
\newblock {\em arXiv preprint arXiv:2104.01318}, 2021.

\bibitem{yu2015multi}
Fisher Yu and Vladlen Koltun.
\newblock Multi-scale context aggregation by dilated convolutions.
\newblock {\em arXiv preprint arXiv:1511.07122}, 2015.

\bibitem{yuan2023small_cfi}
Xiang Yuan, Gong Cheng, Kebing Yan, Qinghua Zeng, and Junwei Han.
\newblock Small object detection via coarse-to-fine proposal generation and imitation learning.
\newblock In {\em Proceedings of the IEEE/CVF international conference on computer vision}, pages 6317--6327, 2023.

\bibitem{zhang2022dino}
Hao Zhang, Feng Li, Shilong Liu, Lei Zhang, Hang Su, Jun Zhu, Lionel Ni, and Heung-Yeung Shum.
\newblock Dino: Detr with improved denoising anchor boxes for end-to-end object detection.
\newblock In {\em The Eleventh International Conference on Learning Representations}, 2022.

\bibitem{zhang2018single}
Shifeng Zhang, Longyin Wen, Xiao Bian, Zhen Lei, and Stan~Z Li.
\newblock Single-shot refinement neural network for object detection.
\newblock In {\em Proceedings of the IEEE conference on computer vision and pattern recognition}, pages 4203--4212, 2018.

\bibitem{zhang2019widerperson}
Shifeng Zhang, Yiliang Xie, Jun Wan, Hansheng Xia, Stan~Z Li, and Guodong Guo.
\newblock Widerperson: A diverse dataset for dense pedestrian detection in the wild.
\newblock {\em IEEE Transactions on Multimedia}, 22(2):380--393, 2019.

\bibitem{zheng2023less}
Dehua Zheng, Wenhui Dong, Hailin Hu, Xinghao Chen, and Yunhe Wang.
\newblock Less is more: Focus attention for efficient detr.
\newblock In {\em Proceedings of the IEEE/CVF International Conference on Computer Vision}, pages 6674--6683, 2023.

\bibitem{zhou2016learning}
Bolei Zhou, Aditya Khosla, Agata Lapedriza, Aude Oliva, and Antonio Torralba.
\newblock Learning deep features for discriminative localization.
\newblock In {\em Proceedings of the IEEE conference on computer vision and pattern recognition}, pages 2921--2929, 2016.

\bibitem{zhu2021detection}
Pengfei Zhu, Longyin Wen, Dawei Du, Xiao Bian, Heng Fan, Qinghua Hu, and Haibin Ling.
\newblock Detection and tracking meet drones challenge.
\newblock {\em IEEE Transactions on Pattern Analysis and Machine Intelligence}, 44(11):7380--7399, 2021.

\bibitem{zhu2020deformable}
Xizhou Zhu, Weijie Su, Lewei Lu, Bin Li, Xiaogang Wang, and Jifeng Dai.
\newblock Deformable detr: Deformable transformers for end-to-end object detection.
\newblock In {\em International Conference on Learning Representations}, 2020.

\bibitem{zong2023detrs}
Zhuofan Zong, Guanglu Song, and Yu Liu.
\newblock Detrs with collaborative hybrid assignments training.
\newblock In {\em Proceedings of the IEEE/CVF international conference on computer vision}, pages 6748--6758, 2023.

\bibitem{zoph2020learning}
Barret Zoph, Ekin~D Cubuk, Golnaz Ghiasi, Tsung-Yi Lin, Jonathon Shlens, and Quoc~V Le.
\newblock Learning data augmentation strategies for object detection.
\newblock In {\em Computer Vision--ECCV 2020: 16th European Conference, Glasgow, UK, August 23--28, 2020, Proceedings, Part XXVII 16}, pages 566--583. Springer, 2020.

\bibitem{zou2023object}
Zhengxia Zou, Keyan Chen, Zhenwei Shi, Yuhong Guo, and Jieping Ye.
\newblock Object detection in 20 years: A survey.
\newblock {\em Proceedings of the IEEE}, 111(3):257--276, 2023.

\end{thebibliography}
}

\newpage
{\appendix[Methodology of CLAP]

\paragraph{Issue of Fixed Image Size for the Deep MLP}

In the main paper, we addressed the challenge commonly faced by most deep MLP models that they only accept input images of a fixed size. This limitation restricts their application in dense prediction tasks, such as object detection, where the input sizes may vary. 
To overcome this limitation, we propose a general Cropping with Adaptive overLaPping (CLAP) approach, aiming at enabling deep MLP models to process images of arbitrary sizes. Subsequently, we assess the effectiveness of this method with the Strip-MLP model in Cross-DINO.

\paragraph{Details of CLAP Approach}

Most MLP-based models only accept input images with fixed size, as the MLP layers are operated on the spatial dimension of the image, resulting in the model’s weights are related to the size of input images.
To address this concern, we propose a simple yet effective CLAP method that involves cropping and padding the image into mini-patches and applying the shared weights to all patches. This allows us to process images of arbitrary sizes using fixed feature weights of deep MLP models, greatly extending their applications on dense prediction tasks.

Initially, we assume that a pre-trained MLP model, such as Strip-MLP model, is available for image classification with a fixed input size of $H_o \times W_o$.
For the downstream task such as object detection, where input images have varying sizes of H × W, we crop the image into mini-patches with overlap, ensuring each patch has the same size as $H_o \times W_o$.
The purpose of splitting the image into mini-patches with overlap is to facilitate information exchange between the patch and neighbouring patches.
As illustrated in Fig.~\ref{fig:mini}, the overlap sizes $l_w$ and $l_h$ are dynamically determined based on the number of neurons in the MLP layer and the size of the images being processed. This process can be formulated as follows:
\begin{figure}[h]
\centering
\includegraphics[width=0.40\textwidth]{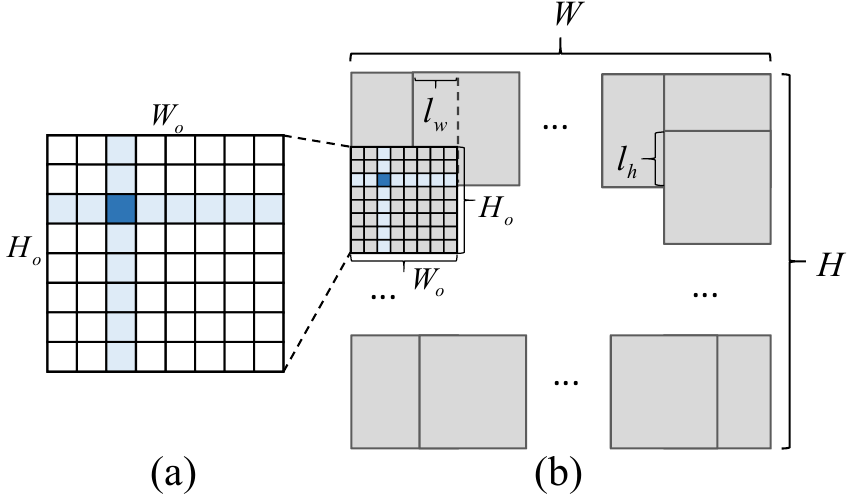} %
\caption{Solution to accept input images with arbitrary sizes for deep MLP models. (a) MLP layers are applied to both the height and width direction of spatial dimension with fixed input size of $H_o \times W_o$. The size of input image is equal to the number of neurons in the MLP layers. (b) The image with the varying size of $H \times W$ is cropped or padded into mini-patches with overlap $l_w$ or $l_h$. 
The mini-patch size is $H_o \times W_o$, which is same as the size of image in (a). 
All mini-patches are then processed by the MLP layers with the shared weight.}
\label{fig:mini}
\end{figure}
\begin{equation}
   n_w = \left[ \frac{W}{W_o} + 0.5 \right]
   \label{eq:n_w}
 \end{equation}
\begin{equation}
   n_h = \left[ \frac{H}{H_o} + 0.5 \right]
   \label{eq:n_h}
 \end{equation}
 \begin{equation}
   l_w = \left[ \frac{(n_w \times W_o -W)}{n_w - 1.5} \right]
   \label{eq:l_w}
 \end{equation}
 \begin{equation}
    l_h = \left[ \frac{(n_h \times H_o -H)}{n_h - 1.5} \right]
   \label{eq:l_h}
 \end{equation}
 where $n_w$ and $n_h$ represent the number of mini-patches split. $H \times W$ denotes the arbitrary size of images being processed. $H_o$ and $W_o$ represent the fixed number of neurons in the MLP layer. The operator $\left[ ~ \right] $ is the floor function symbol.
Specifically, the number of mini-patches, represented by $n_w$ and $n_h$, is determined dynamically based on the ratios of $W$ to $W_o$ and $H$ to $H_o$.
To handle \textit{non-divisible} dimensions, we add $0.5$ to the division results and perform rounding operations, as given in Eq.~\ref{eq:n_w} and Eq.~\ref{eq:n_h}. This ensures that the entire image is processed by dividing it into a greater number of mini-patches.

However, during this process, it may be necessary to pad new image regions with the size of $(n_w \times W_o -W)$ and $(n_h \times H_o -H)$ to accommodate the creation of additional patches.
To address this, instead of padding these regions, our method employs a \textit{cross-region} splitting approach with overlap sizes of $l_w$ and $l_h$, as calculated by Eq.~\ref{eq:l_w} and Eq.~\ref{eq:l_h}. This allows for information exchange between adjacent patches that have overlapping regions, and the overlap region would be processed twice in different patches.

 After applying the MLP layer to all mini-patches, we get the updated features. To restore the mini-patches to their original size, we combine these features of the overlapping regions by summing them with an average weight of 0.5. This ensures that each overlapping region contributes equally to the final restored image.
As a result, we obtain the processed features from the MLP layers, which have the same size as the original features. 
 In summary, our CLAP approach serves as a general method for converting MLP-based models into resolution-free models, enabling their applicability to a wider range of dense prediction tasks.

{\appendix[{Visualization and Analysis}]

\paragraph{Visualization on COCO2017}

We have demonstrated the effectiveness of our Cross-DINO by conducting comparison with other popular advanced models on the COCO2017 dataset in Table~\ref{tab:sota}, clearly confirming the the superior performance of our approach. 
Moreover, the ablation results in Table~\ref{tab:cctm_boost} offer additional evidence supporting the effectiveness of CCTM and Boost loss. These results demonstrate overall consistent improvements across three different backbone models, highlighting the robustness and general effectiveness of our method.

To further visually show the valuable contribution of our method on feature representations for SOD, we visualize the detection results and class activation maps (CAMs)~\cite{zhou2016learning} of different models. We compare the results between the DINO model and the DINO-CB model (DINO with CCTM and Boost Loss, noted as DINO-CB). Using the features and queries from the shallow layers of the encoder features, which typically have a higher resolution, we visualize their detection results and CAM in Fig.~\ref{fig:cam}. 
By comparing the CAMs of the two models, as shown in Fig.~\ref{fig:cam} (b) and (d), we observe that DINO-CB exhibits a higher response on each small object (illustrated by the red box in Fig.~\ref{fig:cam} (d)), indicating that CCTM and Boost Loss help the model pay more attention to small objects and achieve higher performance on SOD. However, the DINO model tends to have higher responses on non-object regions (indicated by the white box in Fig.~\ref{fig:cam} (b)), resulting in lower overall detection performance compared to DINO-CB.
These results from both quantitative and qualitative perspectives provide evidence that our method effectively enhances the detection capability of the detection model specifically for small object detection.

\begin{figure*}[ht]
  \centering
   \includegraphics[width=0.88\linewidth]{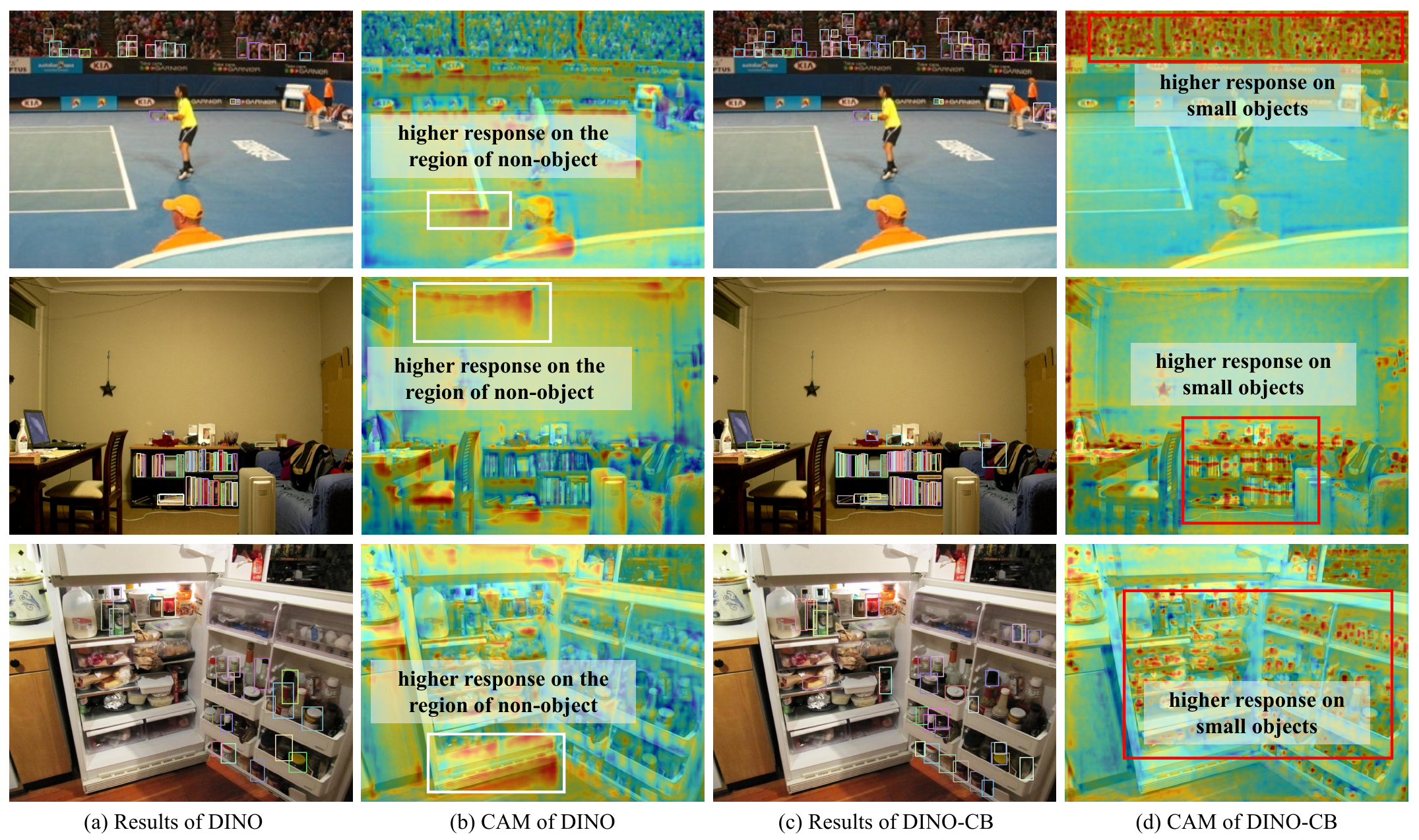}
   \caption{The visualization of detection results of DINO-ResNet50~\cite{zhang2022dino} and DINO-ResNet50 with our CCTM and Boost Loss (noted as DINO-CB) on \texttt{val2017} of COCO~\cite{lin2014microsoft} dataset. (a) and (b) are the detection results of the models. (c) and (d) show the CAMs for these two models.
   }
   \label{fig:cam}
\end{figure*}

\paragraph{Visualization on WiderPerson}

The WiderPerson~\cite{zhang2019widerperson} dataset is known for its challenges in small object detection, as it contains densely populated pedestrians with various occlusions. In our study, we conduct experiments on this dataset to demonstrate the effectiveness of Cross-DINO, as presented in Table~\ref{tab:widerperson}. 
Fig.~\ref{fig:vis_od_acs} provides visualizations of the detection results obtained from both the DINO and Cross-DINO models. These results are based on validation images from the WiderPerson dataset.
By comparing the detection results of the two models, we consistently find that the number of objects (NOB) detected by Cross-DINO is usually higher than that of DINO.
Notably, Cross-DINO surpasses DINO in detecting more challenging objects characterized by occlusions, small sizes, and blurriness. This showcases the robustness and strength of Cross-DINO in detecting `Hard' objects,  ultimately improving the overall detection rate. Additionally, the average confidence score (ACS) of Cross-DINO is significantly higher than that of DINO. This indicates that Cross-DINO effectively aggregates contextual information for detected objects, leading to enhanced confidence in class prediction.

\begin{figure*}[!t]
  \centering
   \includegraphics[width=0.88\linewidth]{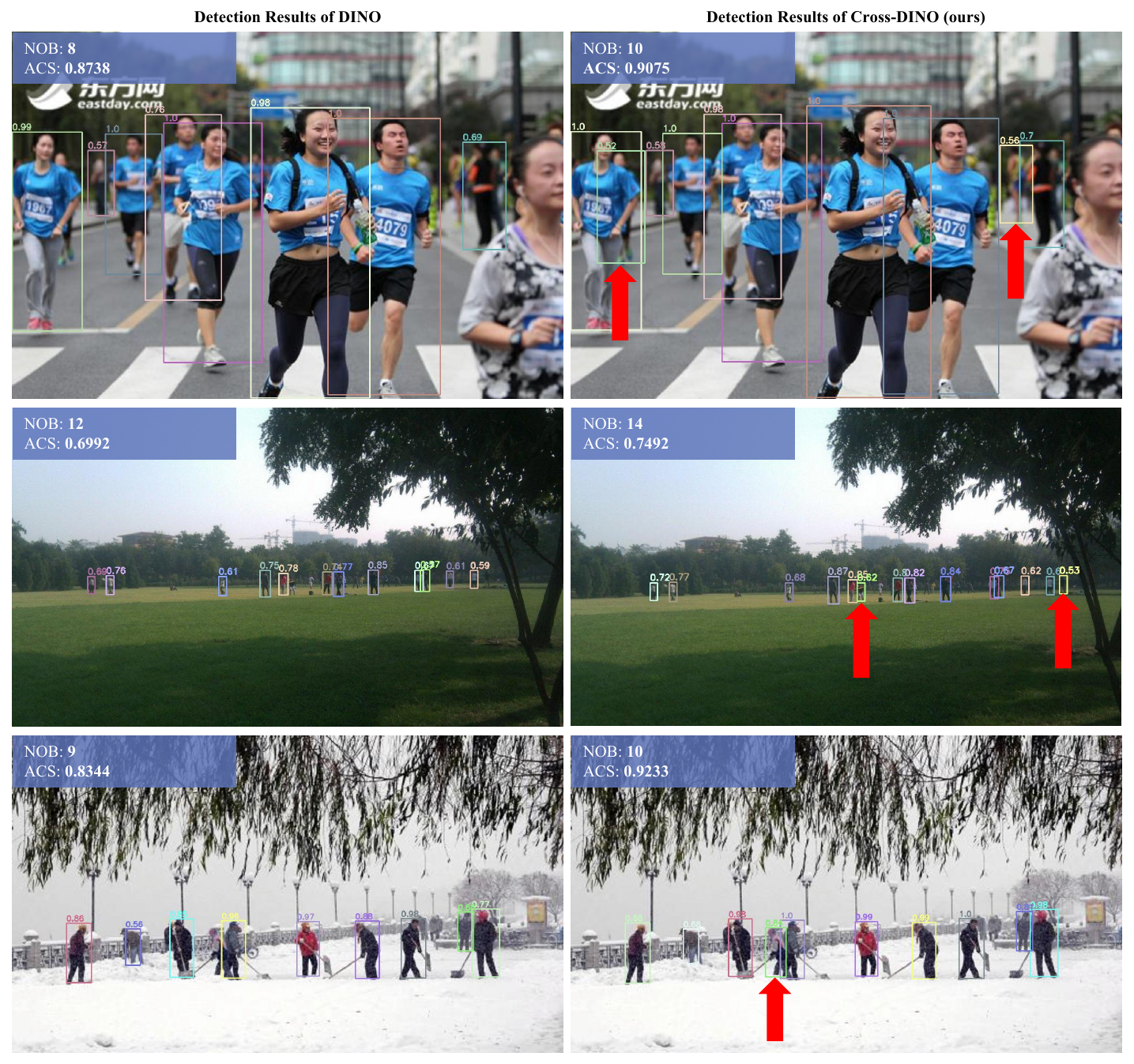}
   \caption{The visualization of detection results of DINO~\cite{zhang2022dino} and Cross-DINO on validation images from the WiderPerson~\cite{zhang2019widerperson} dataset. The visualization includes the number of objects detected (NOB) by each model and the average confidence score (ACS) of the model's class predictions. 
   To ensure a fair comparison of ACS, it is computed only for the objects detected by both DINO and Cross-DINO. Objects that were detected by Cross-DINO but missed by DINO are indicated by a red arrow.
   }
   \label{fig:vis_od_acs}
\end{figure*}

{\appendix[Visualization of Effective Receptive Fields]

To enhance the initial feature representations, we introduce a new CLAP approach that transforms a fixed-resolution deep MLP model into a
flexible-resolution model, termed CLAP-Strip-MLP. This model is capable of processing images of arbitrary sizes without requiring retraining and efficiently aggregates both long-range and short-range information simultaneously. To demonstrate its aggregation capabilities, we visualize the Effective Receptive Fields (ERF)~\cite{luo2016understanding}, which represents the region in the input space that contributes to the activation of a certain output unit. Following the visualization approach used in VMamba~\cite{liu2024vmamba}, we perform a comparative analysis of the ERF for the central pixel across various visual backbones, including ResNet50~\cite{he2016deep}, ConvNeXt-T~\cite{liu2022convnet}, Swin-T~\cite{liu2021swin}, DeiT-S~\cite{touvron2021training}, and VMamba-T~\cite{liu2024vmamba}. We examine the results both before and after training.

As illustrated in Fig.~\ref{fig:erf_strip_mlp}, only DeiT-S, VMamba-T, and our CLAP-Strip-T demonstrate global ERFs, whereas the others exhibit local ERFs. In particular, our CLAP-Strip-T is capable of capturing both global and local ERFs simultaneously, indicating that it can gather richer contextual information.

\begin{figure*}[t]
  \centering
   \includegraphics[width=0.88\linewidth]{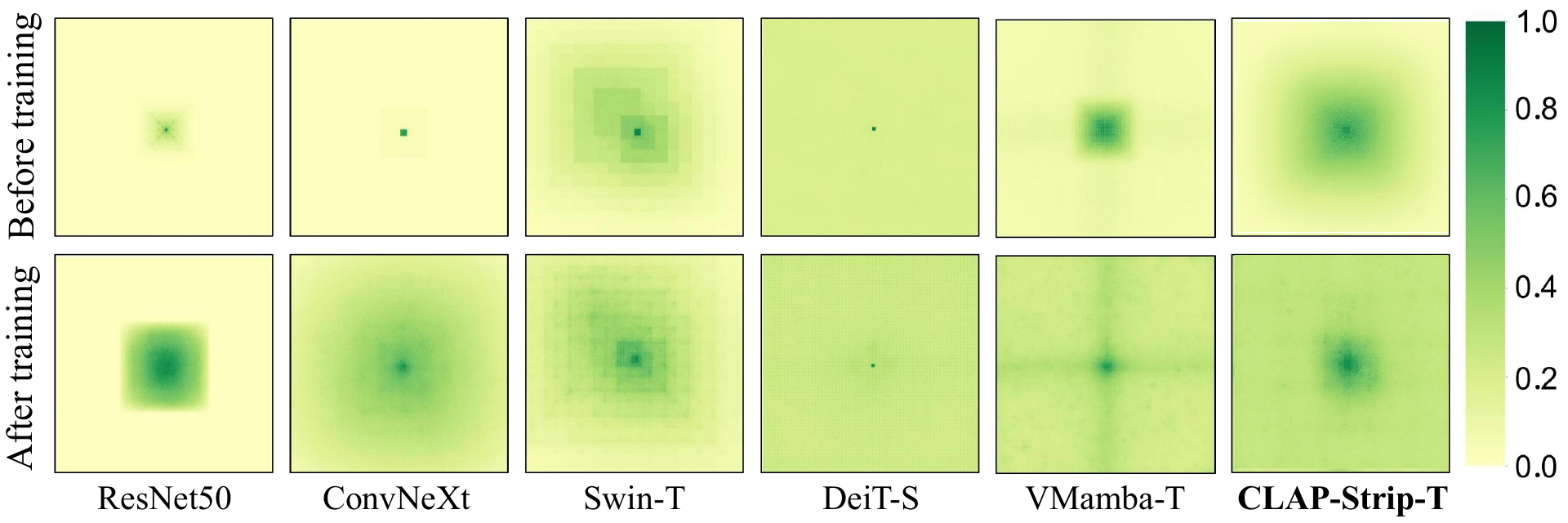}
   \caption{Visualization of Effective Receptive Fields (ERF) between our CLAP-Strip-T and other benchmark models. Pixels with higher intensity signify stronger responses related to the central pixel.}
   \label{fig:erf_strip_mlp}
\end{figure*}

{\appendix[Discussion on the Impacts of $\beta$ in Boost loss]

In Sec.\ref{sec:boost}, we introduce a new scaling factor of $\beta$, where $\beta \in [0, 1]$, to balance the loss weight distributions for objects of different sizes. 
Specifically, this scaling factor affects the loss weight distribution for different object sizes, which in turn influences the model's attention on objects of varying sizes during training, ultimately influencing its accuracy.

To clearly demonstrate the impact of $\beta$ on the loss weights for objects of different sizes, we conducted a weight calculation, as shown in Table~\ref{tab:beta}. The \textit{Relative Distance} (RD) of the loss weights is calculated by dividing the difference in loss weights by the minimum loss weight value. This metric characterizes the degree of attention that different loss functions give to objects of various sizes. A larger RD value indicates a higher level of distinction and focus from the model to objects of different sizes.

For example, we analyzed very tiny objects ($2 \times 2$ pixels) and tiny objects ($8 \times 8$ pixels). Table~\ref{tab:beta} lists the RD values for various $\beta$ values. It can be observed that \textit{a smaller $\beta$ results in a more significant distinction in weights between objects}. Specifically, when $\beta=0.05$, the RD value is scaled up, increasing by \textbf{36.8} times (0.0552 vs. 0.0015) compared to the original loss (i.e., when $\beta=1.0$). A similar phenomenon is noted for both small and large objects, such as objects with $8 \times 8$ and $80 \times 80$ pixels.

\begin{table}[h]
\small
\centering
\caption{Impacts of $\beta$ on loss weights for different object sizes. $\gamma$ is 0.25 in Boost loss. An image size of $1024 \times 1024$ is adopted for calculating $\hat{cs}$.}
\resizebox{1.0\columnwidth}{!}{
\setlength{\tabcolsep}{0.5mm}{
 \renewcommand\arraystretch{1.0}
\begin{tabular}{ c | c | c | c c c }
    \toprule
    Object Size & $\hat{cs}$ & (1-$\hat{cs})^\gamma$ & (1-$\hat{cs}^{0.05})^\gamma$ & (1-$\hat{cs}^{0.1})^\gamma$ & (1-$\hat{cs}^{0.25})^\gamma$ \\
    \midrule
   
    2 $\times$ 2 & 0.0020 & 0.9995 & 0.7189 & 0.8248 & 0.9423\\ 
    8 $\times$ 8 & 0.0078 & 0.9980 & 0.6813 & 0.7874 & 0.9156 \\   
    Relative Distance & - & 0.0015 & 0.0552 (\textbf{36.8}$\uparrow$) & 0.0475(31.7$\uparrow$) & 0.0292(19.5$\uparrow$) \\ 
    \midrule
    8 $\times$ 8 & 0.0078 & 0.9980 & 0.6813 & 0.7874 & 0.9156\\ 
    80 $\times$ 80 & 0.0781 & 0.9799 & 0.5882 & 0.6888 & 0.8286 \\   
    Relative Distance & - & 0.0185 & 0.1583 (\textbf{8.6}$\uparrow$) & 0.1431 (7.7$\uparrow$) & 0.1050 (5.7$\uparrow$) \\ 
     \bottomrule
     \end{tabular}}
     }
\label{tab:beta}
\end{table}

To identify the optimal value of $\beta$ for our model, we conducted additional ablation experiments on the scaling factor $\beta$. As illustrated in Table~\ref{tab:ablation_scale}, different values of $\beta$ result in varying levels of overall enhancement in model accuracy. In particular, Cross-DINO achieved increases of \textbf{+1.4\%} $\sim$ \textbf{+2.0\%} in AP on the VisDrone2019 test dataset, demonstrating the robustness of the model across different $\beta$ values.  
As discussed in Table~\ref{tab:beta}, the value of $\beta$ impacts the distribution of loss weights for objects of different sizes. Notable, when compared to $\beta=0.05$, the model with $\beta=0.25$ gets higher AP (35.8\% vs. 35.2\%) but with a little performance drop in AP$_{vt}$ (with smaller RD for very tiny objects using $\beta=0.25$ than using $\beta=0.05$), highlighting the influence of $\beta$ on objects with different sizes. Based on these findings, we set $\beta = 0.1$ for the main results presented in Table~\ref{tab:visdrone}.

\begin{table}[h]
\small
\centering
\caption{Ablation experiments on the scaling factor $\beta$ of Boost loss on the VisDrone2019 test dataset.}
\resizebox{1.0\columnwidth}{!}{
\setlength{\tabcolsep}{1.0mm}{
 \renewcommand\arraystretch{1.0}
\begin{tabular}{ c | c | c | c | c c c c c c }
    \toprule
    Model & Eps & Loss & $\beta$ & AP & AP$_{0.5}$ & AP$_{0.75}$ & AP$_{vt}$ & AP$_{t}$ & AP$_{s}$ \\
    \midrule
    
    Cross-DINO-CLAP & 12 & Focal & - & 33.8 & 56.1 & 34.6 & 8.4 & 15.9 & 29.5 \\
    Cross-DINO-CLAP & 12 & Boost & 0.05 & 35.2 & 59.4 & 35.9 & 8.3 & 16.1 & 30.6 \\
    Cross-DINO-CLAP & 12 & Boost & 0.10 & 35.4 & 59.8 & 35.9 & 8.2 & 16.2 & 31.5 \\
    Cross-DINO-CLAP & 12 & Boost & 0.25 & 35.8 & 59.8 & 36.4 & 7.6 & 16.5 & 31.1 \\
    
     \bottomrule
     \end{tabular}}
     }
\label{tab:ablation_scale}
\end{table}

{\appendix[Differences between Boost loss and focal loss]

The \textbf{\textit{motivation}}, \textbf{\textit{prior knowledge}}, \textit{\textbf{implementation}}, and \textit{\textbf{performance}} are fundamentally different between our Boost loss and existing Focal loss~\cite{lin2017focal}. Specifically, we detail their differences as following:

\begin{table}[t]
\small
\centering
\caption{Ablation results of Focal loss and Boost loss on VisDrone2019 test dataset.}
\resizebox{1.0\columnwidth}{!}{
\setlength{\tabcolsep}{1.0mm}{
 \renewcommand\arraystretch{1.0}
\begin{tabular}{ c | c | c | c c c c c c }
    \toprule
    Model & Epoch & Loss & AP & AP$_{0.5}$ & AP$_{0.75}$ & AP$_{vt}$ & AP$_{t}$ & AP$_{s}$ \\
    \midrule
    DINO-R50 & 12 & Focal & 31.7 & 54.1 & 31.9 & 5.9 & 14.0 & 27.4 \\ 
    DINO-R50 & 12 & Boost & 32.8 & 56.2 & 32.9 & 5.6 & 14.4 & 28.4 \\ 
    \midrule
    Cross-DINO-R50 & 12 & Focal & 32.1 & 54.1 & 32.5 & 5.1 & 13.6 & 27.7 \\  
    Cross-DINO-R50 & 12 & Boost & 33.1 & 55.8 & 33.7 & 5.7 & 14.5 & 28.5  \\  

    \midrule
    Cross-DINO-CLAP & 12 & Focal & 33.8 & 56.1 & 
    34.6 & 8.4 & 15.9 & 29.5 \\ 
    Cross-DINO-CLAP & 12 & Boost & 35.2 & 59.4 & 35.9 & 8.3 & 16.1 & 30.6 \\  
     \bottomrule
     \end{tabular}}
     }
\label{tab:ablation_loss}
\end{table}

\begin{itemize}
    \item \textbf{Different Motivations.} Boost loss is specifically designed to address the significant correlations between the \textit{confidence} and the \textit{box size} of small and large objects, as illustrated in Fig.~\ref{fig:class_pro} and Sec.~\ref{sec:boost} in the main text. In contrast, Focal loss~\cite{lin2017focal} primarily aims to address the issue of \textit{class imbalance} across different classes. These differing motivations allow each loss function to effectively address specific challenges.
    
    \item \textbf{Different Priors.} Boost loss works by integrating \textit{\textbf{additional}} prior information regarding \textit{\textbf{box sizes}} in training process. It automatically identifies smaller objects as hard samples based on their ground truth box size annotations within the image. By explicitly assigning higher weights to small objects, Boost loss prioritizes these hard samples, thereby focusing the training on them. In contrast, \textit{Focal loss does not leverage this box size information}, which diminishes its ability for recognizing small objects.
    
    \item \textbf{Different Implementations}. Compared to Focal loss, our Boost loss focuses on modeling the consistent relation between the predicted object size and confidence by defining a \textit{\textbf{new ground truth}} termed CS, which is built on the existing label annotation information of class and box (without the need for additional labeling costs), as outlined in Sec.~\ref{sec:boost} of the ``Method'' section.
    
    \item \textbf{Performance Differences}. The model trained with Boost loss \textit{\textbf{consistently outperforms}} the model trained with Focal loss, across \textit{\textbf{various backbones}} (ResNet50, Swin-T, CLAP-Strip-T) and \textit{\textbf{datasets}} (COCO and VisDrone2019). For COCO, DINO with Boost loss achieves improved performance with increases of \textbf{+0.6\%/+0.6\%/+0.9\%} in AP and \textbf{+0.2\%/+1.0\%/+1.2\%} in AP$_S$ across the three backbone models, as shown in Table~\ref{tab:cctm_boost} of the main text, demonstrating its general effectiveness. Furthermore, we also conducted \textit{additional} experiments on the more challenging SOD dataset, VisDrone2019~\cite{zhu2021detection}. As presented in Table~\ref{tab:ablation_loss}, both DINO and Cross-DINO models trained with Boost loss \textit{\textbf{outperform}} those trained with Focal loss. Specifically, we observed increases of \textbf{+1.1\%} AP for DINO-R50, \textbf{+1.0\%} AP for Cross-DINO-R50, and \textbf{+1.4\%} AP for Cross-DINO-CLAP-Strip-T models, further confirming its effectiveness.
\end{itemize}

\vfill

\end{document}